\pdfoutput=1

\documentclass[11pt]{article}

\usepackage[final]{acl}

\usepackage{times}
\usepackage{latexsym}
\usepackage{multirow}
\usepackage{booktabs}
\usepackage{algorithm}
\usepackage{algpseudocode}
\usepackage{amsmath}
\usepackage{array}
\usepackage{makecell}

\usepackage{tcolorbox}
\usepackage{soul}
\usepackage{xcolor}
\usepackage[T1]{fontenc}

\usepackage[utf8]{inputenc}

\usepackage{microtype}

\usepackage{inconsolata}

\usepackage{graphicx}
\usepackage{hyperref}
\usepackage{enumitem}
\usepackage{listings}
\usepackage{amssymb}
\usepackage{geometry}
\usepackage{colortbl}
\usepackage{arydshln}
\usepackage{textcomp}
\usepackage{tabularx}

\title{\emph{TreeReview}: A Dynamic Tree of Questions Framework for Deep and Efficient LLM-based Scientific Peer Review}

\makeatletter
\def\@fnsymbol#1{}
\makeatother

\author{
 \textbf{Yuan Chang\textsuperscript{1,2 *}}\thanks{*\ Equal contribution.},
 \textbf{Ziyue Li\textsuperscript{1,2 *}},
 \textbf{Hengyuan Zhang\textsuperscript{3 \dag}} \thanks{\dag\ Corresponding author.}, 
 \textbf{Yuanbo Kong\textsuperscript{1,2}},
 \textbf{Yanru Wu\textsuperscript{4}},
 \\
 \textbf{Hayden Kwok-Hay So\textsuperscript{3}},
 \textbf{Zhijiang Guo\textsuperscript{5 \dag}},
 \textbf{Liya Zhu\textsuperscript{1,2 \dag}},
 \textbf{Ngai Wong\textsuperscript{3}}
\\
 \textsuperscript{1}National Science Library, Chinese Academy of Sciences \\
 \textsuperscript{2}Department of Information Resources Management, School of Economics and Management, \\
 University of Chinese Academy of Sciences 
 \textsuperscript{3}The University of Hong Kong\\
 \textsuperscript{4}Tsinghua University
 \textsuperscript{5}Hong Kong University of Science and Technology (Guangzhou)
 \\
\texttt{\{\hypersetup{urlcolor=black}\href{mailto:changyuan@mail.las.ac.cn}{changyuan}, \href{mailto:liziyue@mail.las.ac.cn}{liziyue}, \href{mailto:zhuly@mail.las.ac.cn}{zhuly}\}@mail.las.ac.cn \ \  hengyuan.zhang88@gmail.com}  
}
\definecolor{lightyellow}{rgb}{1, 1, 0.8}
\sethlcolor{lightyellow}

\begin{document}
\maketitle
\begin{abstract}
While Large Language Models (LLMs) have shown significant potential in assisting peer review, current methods often struggle to generate thorough and insightful reviews while maintaining efficiency. In this paper, we propose \emph{TreeReview}, a novel framework that models paper review as a hierarchical and bidirectional question-answering process. \emph{TreeReview} first constructs a tree of review questions by recursively decomposing high-level questions into fine-grained sub-questions and then resolves the question tree by iteratively aggregating answers from leaf to root to get the final review. Crucially, we incorporate a dynamic question expansion mechanism to enable deeper probing by generating follow-up questions when needed. We construct a benchmark derived from ICLR and NeurIPS venues to evaluate our method on full review generation and actionable feedback comments generation tasks. Experimental results of both LLM-based and human evaluation show that \emph{TreeReview} outperforms strong baselines in providing comprehensive, in-depth, and expert-aligned review feedback, while reducing LLM token usage by up to 80\% compared to computationally intensive approaches. Our code and benchmark dataset are available at \url{https://github.com/YuanChang98/tree-review}.
\end{abstract}

\section{Introduction}
The exponential growth in academic publications has placed increasing strain on the peer review system, which remains the primary quality control mechanism for scientific research \cite{larsen2010rate, gropp2017peer}. The widening gap between submission volume and reviewer availability has led to bottlenecks that potentially delay scientific progress \cite{leopold2015increased}. Thus, there is an urgent need for automated methods to support the peer review process, which can provide assistance to reviewers and help authors improve their manuscripts, maintaining the quality and efficiency of scholarly evaluation \cite{lin2023automated}.

\begin{figure}[t]
    \centering
    \includegraphics[width=\columnwidth]{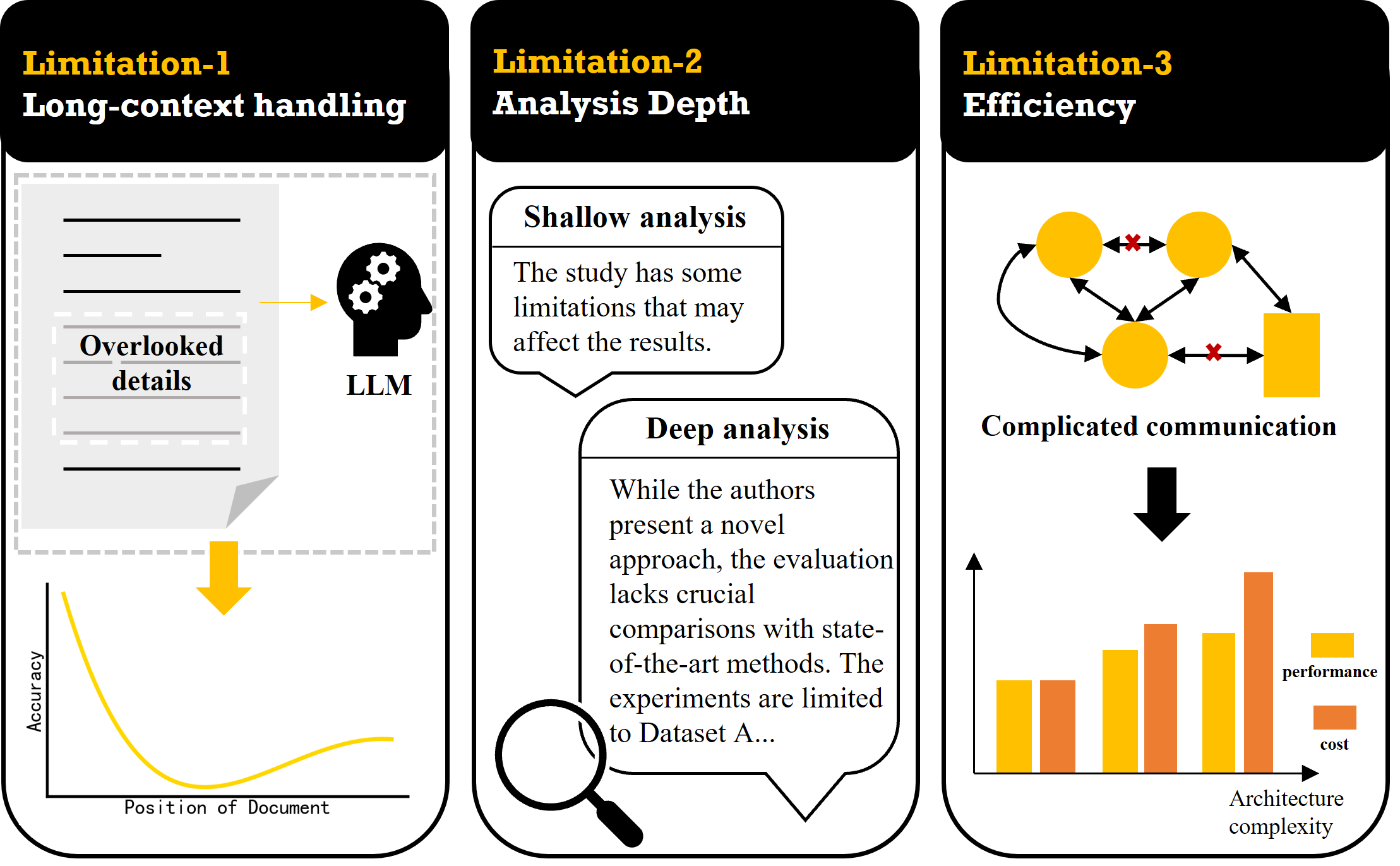}
    \caption{Current LLM-driven review methods face key limitations in: handling long papers, providing deep analysis, and managing computational costs. }
    \label{fig:limitation}
\end{figure}

Large Language Models (LLMs) have demonstrated remarkable capabilities across a wide range of scientific tasks \cite{zheng2023large, wang2024autosurvey, LiSurvey2025} and have also been increasingly applied to assist in scholarly peer review \cite{ZHUANG2025103332}. 
Recent studies have utilized elaborate prompting strategies \cite{liang2024can}, fine-tuned models \cite{yu-etal-2024-automated,gao2024reviewer2}, and multi-agent frameworks \cite{d2024marg} in attempts to replicate expert-level peer review procedures.

While these advances have shown promise in generating feedback for scientific papers, three critical limitations (shown in Fig.~\ref{fig:limitation}) hinder their real-world application. First, despite significant advances in LLMs' ability to process inputs spanning millions of tokens \cite{zhou2024survey, liu2025comprehensive}, recent studies reveal persistent challenges of LLMs in capturing long-range dependencies \cite{li-etal-2024-loogle}, attending to information located mid-context \cite{liu-etal-2024-lost,liu2025mudaf}, and reasoning over complex inputs \cite{li2024long,tong2024can}. Scientific papers present particular difficulties due to their lengthy nature, with technical details dispersed throughout the paper. As a result, important and fine-grained details can be overlooked, leading to incomplete reviews. Second, these methods often produce superficial feedback, lacking the depth required to critically evaluate a paper's technical nuances \cite{zhou-etal-2024-llm,du-etal-2024-llms,liang2024can}. Finally, while multi-agent frameworks such as MARG \cite{d2024marg} achieve strong performance, their sophisticated design requires extensive interaction and coordination between agents, leading to substantial computational overhead and vulnerability to communication errors.

In this work, we propose \emph{TreeReview}, a dynamic tree of questions framework that structures LLM-based peer review as a hierarchical, question-driven reasoning process to efficiently generate in-depth feedback for lengthy papers. \emph{TreeReview} tackles the identified challenges through the following design: 1) To avoid overlooking paper details, it decomposes the high-level review task into a tree of fine-grained review questions and answers them using focused, relevant paper chunks; 2) To overcome superficial feedback, it recursively refines broad review aspects into specific inquiries and employs a dynamic question expansion mechanism for deeper, context-aware probing; 3) It leverages explicit and structured decomposition and aggregation strategy to avoid complex multi-agent interactions, thereby minimizing token usage. Operationally, \emph{TreeReview} functions in two stages: 1) a \textbf{Top-Down} stage, where broad review questions are recursively decomposed into specific sub-questions forming a review question tree; 2) a \textbf{Bottom-Up} stage, where answers are aggregated from leaf to root to synthesize comprehensive feedback, with dynamic expansions for deeper investigation when needed.

To systematically evaluate our framework, we construct a diverse benchmark comprising papers and human reviews from ICLR and NeurIPS venues, enabling both full review generation and actionable feedback comments generation assessment. Extensive experiments demonstrate the effectiveness and efficiency of \emph{TreeReview}. For full review generation task, results show that \emph{TreeReview} outperforms baselines in LLM-as-Judge evaluation, achieving the highest score across critical quality dimensions such as specificity (\textuparrow 12.27\% over the best baseline), comprehensiveness (\textuparrow 11.22\%), and technical depth (\textuparrow 6.45\%). In the alignment evaluation for feedback comments generation task, \emph{TreeReview} achieves the highest precision and outperforms the strong baseline MARG by 5.7\% in Jaccard while reducing token usage by 80.2\%. Further human evaluation results show that \emph{TreeReview} produces reviews that are more preferred by expert evaluators over baseline methods with high consistency.
Our main contributions are summarized as follows:
\vspace{0pt}
\begin{itemize}[leftmargin=*, topsep=0pt]
\setlength{\itemsep}{0pt}
\setlength{\parsep}{0pt}
\setlength{\parskip}{0pt}
\item We propose \emph{TreeReview}, a novel framework to address key challenges in LLM-based scientific peer review.
\item We construct and open-source an evaluation benchmark for full review generation and actionable feedback comments generation scenarios to facilitate future research.
\item We conduct extensive experiments showing that \emph{TreeReview} outperforms strong baselines in providing high-quality and well-aligned review feedback while maintaining efficiency.
\end{itemize}

\section{Related Work}
\subsection{Automated Scientific Peer Review}

Automated peer review has long been explored to address the increasing burden on traditional review processes. Early efforts focused on specific aspects such as reviewer assignment~\cite{10.1145/3292500.3330899}, plagiarism detection~\cite{10.1145/3345317}, and paper rating recommendation~\cite{kang-etal-2018-dataset, deng-etal-2020-hierarchical}. More recent work has attempted to generate free-form paper reviews using small language models with supervised fine-tuning approaches~\cite{10.1613/jair.1.12862, lin2023moprd}, but these models' limited context length and comprehension capabilities make it challenging to generate nuanced and comprehensive reviews for full-length academic papers.

Recently, LLMs have showcased remarkable performance in several application scenarios, such as reasoning~\citep{li2025system,yu2025chain}, multilingualism~\citep{huang2023not,gurgurov2024multilingual,zhang2024shifcon}, and text generation in specialized contexts~\citep{yang2024llm,yang2025quantifying,liang2024task,TanProxy2024, zhang2024balancing,chang-etal-2024-guiding,li-etal-2024-simulating}.
Leveraging LLMs to assist peer review has recently become an emerging research direction, exploring how LLMs can potentially augment the challenging task of scholarly evaluation \cite{ZHUANG2025103332}.
Several studies have evaluated or benchmarked the performance of state-of-the-art LLMs in generating paper reviews \cite{liang2024can,zhou-etal-2024-llm,zhou2024may,du-etal-2024-llms,mahmoud2024evaluating}. These works demonstrate that while LLMs can provide meaningful feedback, they often struggle with critical analysis and tend to produce reviews that lack the depth and specificity found in human-written reviews \cite{liang2024can, zhou-etal-2024-llm, du-etal-2024-llms}.

Research has progressed along two principal trajectories for enhancing review quality beyond simple prompting. The first involves curating peer review datasets and fine-tuning LLMs specifically for review generation \cite{yu-etal-2024-automated,gao2024reviewer2}. The second direction focuses on more complex frameworks that enhance LLM capabilities through multi-agent systems, multi-modal information processing, and external knowledge integration \cite{d2024marg,wang2024mamorx,chamoun-etal-2024-automated}. Beyond standalone review generation, researchers also explore integrating it into automated scientific discovery frameworks such as the AI Scientist \cite{lu2024ai} and the CycleResearcher \cite{weng2025cycleresearcher} to serve as a crucial feedback module.

\subsection{Decomposition of Complex Tasks}
Task decomposition has been extensively studied in NLP as an effective strategy to address challenging reasoning tasks by dividing them into manageable sub-tasks \cite{perez-etal-2020-unsupervised,guo-etal-2022-complex, zheng2023outline,wang-etal-2023-plan}, particularly for tasks requiring multi-step reasoning and comprehensive analysis. Techniques such as Chain-of-Thought (CoT) \cite{wei2022chain} prompting encourage LLMs to generate intermediate reasoning steps, implicitly decomposing the problem-solving process. Subsequent research further advances this approach by explicitly breaking down problems into discrete sub-problems, which are then solved sequentially or iteratively \cite{DBLP:journals/corr/abs-2210-02406,press-etal-2023-measuring,zhou2023leasttomost,dua-etal-2022-successive}. Furthermore, tree-based reasoning structures \cite{yang2024selfgoal,prasad-etal-2024-adapt,Zhang_Zeng_Meng_Wang_Sun_Bai_Shen_Zhou_2024,zhao-etal-2024-epo,yu2025chain,li2025system} have been proposed to model the hierarchical dependencies within complex tasks, enabling a more comprehensive decomposition and result aggregation process.

The scientific peer review process inherently involves a highly complex cognitive task that demands comprehensive evaluation across multiple aspects. Our work handles it by employing a dynamic hierarchical decomposition of the review task, which enables each aspect of the papers to be assessed in a focused manner. 

\begin{figure*}[t]
    \centering
    \includegraphics[width=\textwidth]{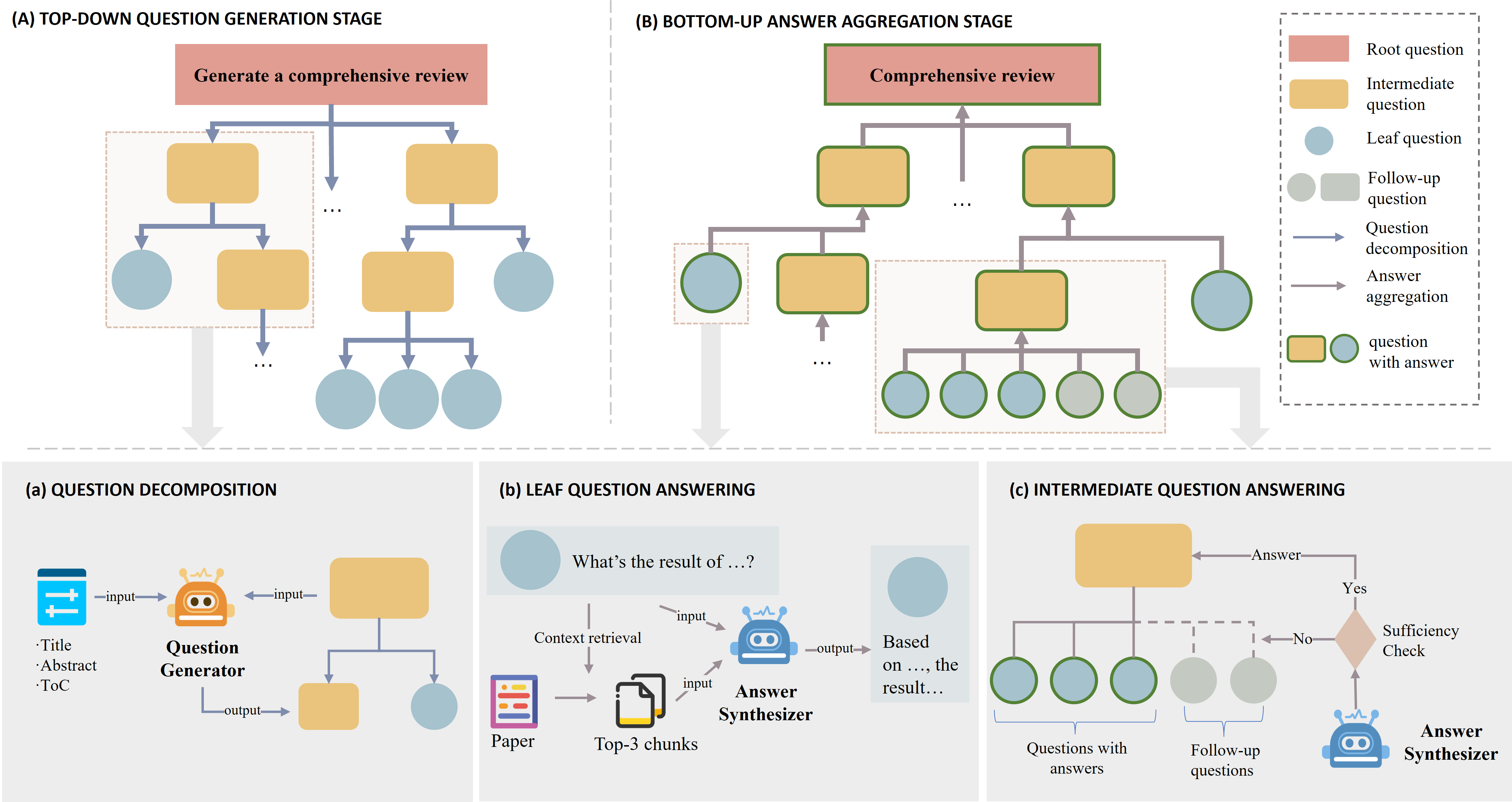}
    \caption{Overview of \emph{TreeReview} framework. \textbf{(A)} Top-down Question Generation Stage: The overall process of hierarchical question decomposition. \textbf{(B)} Bottom-up Answer Aggregation Stage: The overall process of aggregating answers from leaf to root, producing the final review. \textbf{(a)} Decomposing a non-leaf question. \textbf{(b)} Answering a leaf question. \textbf{(c)} Dynamically raising follow-up sub-questions and synthesizing the answer for an intermediate question.}
    \label{fig:framework}
\end{figure*}

\section{Method}

\subsection{Overview}
Human reviewers often conduct a review by first raising exploratory review questions about the paper to guide their reading and then addressing these questions for deeper comprehension.\footnote{This practice aligns to some extent with the reviewer guidelines of \href{https://aclrollingreview.org/reviewerguidelines}{ARR}, \href{https://iclr.cc/Conferences/2024/ReviewerGuide}{ICLR}, \href{https://neurips.cc/Conferences/2024/ReviewerGuidelines}{NeurIPS}, \href{https://plos.org/resource/how-to-read-a-manuscript-as-a-peer-reviewer/}{PLOS}, and \href{https://authorservices.wiley.com/Reviewers/journal-reviewers/how-to-perform-a-peer-review/step-by-step-guide-to-reviewing-a-manuscript.html}{WILEY}, etc.} Inspired by this cognitive pattern, we propose \emph{TreeReview}, a dynamic tree of questions framework to model scientific paper review as a tree-like reasoning process. 

As illustrated in Fig.~\ref{fig:framework}, \emph{TreeReview} includes two stages: \uppercase\expandafter{\romannumeral1}. \textit{Top-down review question generation} stage (§\ref{sec:top_down_generation}), where a question generator agent $M_q$ recursively decomposes high-level review questions into increasingly fine-grained ones, establishing a question tree of exploration; \uppercase\expandafter{\romannumeral2}. \textit{Bottom-up answer aggregation} stage (§\ref{sec:bottom_up_aggregation}), where an answer synthesizer agent $M_a$ iteratively synthesizes answers up the tree to delve into the paper content and make the final review at the root. Crucially, \emph{TreeReview} incorporates a \textit{dynamic review question expansion} mechanism, where $M_a$ can raise follow-up questions based on the current state to probe areas of the paper requiring deeper investigation.

This hierarchical and bidirectional architecture enables a focused and in-depth \textbf{local} analysis of specific paper details often obscured in long contexts, while constructing comprehensive \textbf{global} assessments through systematic aggregation.

\subsection{Top-down Question Generation Stage}
\label{sec:top_down_generation}
For a given paper, we first construct a review question tree $\mathcal{T}$ in a top-down manner. The process begins with the top-level review task (e.g., ``Generate a comprehensive peer review for this paper'') as the root question and recursively decomposes it into increasingly focused sub-questions. As illustrated in Fig.~\ref{fig:framework}(a), for each non-leaf question $q_i$ with depth $l < D_\text{max}$, we employ a specialized \textit{Question Generator} agent $M_{a}$ to decompose it into at most $W_\text{max}$ sub-questions:
\begin{equation}
q_{i,1}, \cdots , q_{i,n} \mid \emptyset = M_q(q_i, \mathcal{P}_\text{meta}, l)
\end{equation}
where $n \leq W_\text{max}$ and $\mathcal{P}_\text{meta}$ represents the metadata (title, abstract, and table of contents) of paper $ \mathcal{P}$. Notably, this decomposition is adaptive: $M_q$ generates more sub-questions for broader questions to ensure coverage, while more specific questions lead to fewer sub-questions. If $M_q$ determines that $q_i$ is sufficiently specific and requires no further decomposition, it returns $\emptyset$, and we mark $q_i$ as a leaf question.
We leverage only the metadata rather than the full paper as the source to encourage $M_{q}$ to generate more exploratory questions without being constrained by localized context. 
\vspace{3pt}

\noindent\textbf{Question Generator Action Principles.} \quad $ M_q $ is implemented as an LLM-based agent guided by a carefully crafted prompt that emphasizes the following principles:

\begin{itemize}[leftmargin=*, topsep=0pt]
\setlength{\itemsep}{0pt}
\setlength{\parsep}{0pt}
\setlength{\parskip}{0pt}
\item The decomposition strategy is depth-aware: At depth 1 (root level), it generates broad questions covering major review aspects (novelty, methodology, significance, etc.), while at deeper levels, it generates increasingly specific questions that probe finer details.
\item All generated sub-questions adhere to the Mutually Exclusive, Collectively Exhaustive (MECE) principle, ensuring that they are non-overlapping and jointly cover the parent question’s scope.
\end{itemize}

\subsection{Bottom-up Answer Aggregation Stage}
\label{sec:bottom_up_aggregation}
In this stage, the review question tree $\mathcal{T}$ is systematically resolved from leaf to root, wherein an answer synthesizer $M_a$ progressively traverses the tree to:
 1) answer leaf questions with contextually relevant paper content, 2) synthesize answers for intermediate questions based on their sub-question answer pairs, and 3) culminate in generating the final review at the root. 
 This bottom-up aggregation process distills fine-grained observations into increasingly higher-level insights, enhancing both depth and comprehensiveness in the review feedback. We describe each type of step in detail below.

\vspace{3pt}

\noindent\textbf{Leaf Question Answering Operation.} \quad Leaf questions in $\mathcal{T}$ focus on specific paper details. Instead of using the full paper as context, which can reduce inference efficiency and potentially distract $M_a$ from the pertinent information, we seek to identify the most relevant content from the paper to serve as the source. To this end, $\mathcal{P}$ is first segmented into chunks of size $L$. For each leaf question $q_i^\text{leaf}$, we utilize the question-aware coarse-grained context compression method from LongLLMLingua \cite{jiang-etal-2024-longllmlingua} to filter out the top-$k$ most relevant chunks, based on the probability of $q_i^\text{leaf}$ conditioned on each chunk $p(q_i^\text{leaf}\mid \text{chunk})$.\footnote{In this work, we use \href{https://huggingface.co/meta-llama/Llama-3.1-8B-Instruct}{Llama-3.1-8B-Instruct} to calculate the probability, more details can be seen in Appendix \ref{appendix:chunk_rerank_details}.} As shown in Fig. \ref{fig:framework}(b), $M_a$ use this focused context to answer $q_i^\text{leaf}$:
\begin{equation}
    a_i = M_a(q_i^\text{leaf}, \{\text{chunk}_{r_1}, \dots, \text{chunk}_{r_k}\})
\end{equation}
where $r_1, \dots, r_k$ are subscripts for top-$k$ most relevant chunks to $q_i^\text{leaf}$.

Additionally, $M_a$ is instructed to ground its answer by explicitly citing evidence from the provided context chunks, which can facilitate the reliability of subsequent answer aggregation steps.

\vspace{3pt}

\noindent\textbf{Answer Aggregation Operation.} \quad
For each intermediate question $q_i^\text{inter}$, i.e. non-leaf and non-root question, the answer $a_i$ is synthesized by aggregating the answers from its sub-questions:
\begin{equation}
    a_i = M_a(q_i^\text{inter}, \{(q_{i,j}, a_{i, j})\}_{j=1}^{n_i})
\end{equation}
where $\{(q_{i,j}, a_{i, j})\}_{j=1}^{n_i}$ represents the set of sub-question and answer pairs for $q_i^\text{inter}$.
Recognizing that this initial set may not always provide sufficient information for comprehensive answer synthesis, we further introduce a \emph{dynamic review question expansion} mechanism that allows deeper exploration of paper content when needed. Specifically, as illustrated in Fig.~\ref{fig:framework}(c), when resolving an intermediate question $q_i^{inter}$,  $M_a$ first evaluates whether the insights and evidence presented in current sub-questions and answers suffice to resolve $q_i^{inter}$. If deemed sufficient, $M_a$ proceeds with synthesis. 
Otherwise, $M_a$ proposes up to $W_{\text{max}}^{\raisebox{0.5ex}{\scriptsize{exp}}}$ follow-up questions $q_{i,\bar{n}_i+1}, \cdots , q_{i,\bar{n}_i+m}$ ($m \leq W_{\text{max}}^{\raisebox{0.5ex}{\scriptsize{exp}}}$) based on the current state (i.e. $q_i^\text{inter}, \{(q_{i,j}, a_{i,j})\}_{j=1}^{\bar{n}_i}$) to probe unaddressed aspects, where $\bar{n}_i$ is the number of current sub-questions. 
These follow-up questions are integrated into the question tree $\mathcal{T}$ and further decomposed (if needed) by $M_q$. The answer synthesis for $q_i^\text{inter}$ is deferred until the answers for all the newly expanded sub-questions are obtained.

Our ablation studies (§\ref{sec:ablation_study}) and case analysis (Appendix \ref{appendix:case_study}) demonstrate that this mechanism can effectively uncover nuanced aspects overlooked by the initial question tree and contribute to identifying critical paper issues.

\vspace{3pt}

\noindent\textbf{Final Review Generation.} \quad
Upon reaching the root review task $q_\text{root}$, all its sub-questions and corresponding answers $\{(q_{\text{root},j}, a_{\text{root},j})\}_{j=1}^{n_\text{root}}$ have been collected. Subsequently, the final review $\mathcal{R}$ is generated. Unlike intermediate aggregation, which relies solely on sub-answers, this final step incorporates the full paper $\mathcal{P}$ to provide holistic context, and the answers for its sub-questions serve as explicit reasoning traces guiding the review process of $M_a$:
\begin{equation}
    \mathcal{R} = M_a(\mathcal{P}, \{(q_{\text{root},j}, a_{\text{root},j})\}_{j=1}^{n_\text{root}}, \text{Inst}_{\mathcal{R}})
\end{equation}
where $\text{Inst}_{\mathcal{R}}$ is the additional instruction for regularizing the review format. 
More implementation details of \emph{TreeReview} are provided in Appendix \ref{appendix:method_detail}.

\section{Experiments}
\subsection{Experimental Settings}
\noindent \textbf{Tasks.} \quad We evaluate our proposed framework on two distinct review scenarios: (1) Full Review Generation: This task involves producing a comprehensive review, including summary, strengths, weaknesses, questions and ratings, mirroring the complete review process of typical academic venues; (2) Actionable Feedback Comments Generation: This task focuses on generating a list of specific, critical feedback points targeting substantive weaknesses and improvement areas in a paper. Such actionable comments are highly valued in real-world peer review for directly helping authors identify and address major flaws, yet they pose unique challenges distinct from holistic review generation, as models must pinpoint and articulate concrete issues rather than summarizing general impressions. We leverage these settings to test \emph{TreeReview} in handling both holistic assessments and targeted critiques. 

\vspace{3pt}

\noindent \textbf{Baselines.} \quad For full review generation, we consider two categories of baselines: 1) Supervised fine-tuning (SFT) methods: \textsc{Reviewer2} \cite{gao2024reviewer2} and SEA-E \cite{yu-etal-2024-automated}, both 7B-parameter models specifically fine-tuned on this task; 2) Prompting-based methods: \textbf{D}irect prompting with step-by-step review \textbf{G}uidelines and few-shot review \textbf{E}xamples (DGE) which we adopt as proxies for the methods of \citet{du-etal-2024-llms} and \citet{lu2024ai}; and the method of \citet{liang2024can}, which we refer to as SORT (\textbf{S}tructured \textbf{O}utline \textbf{R}eview \textbf{T}emplate), that generates reviews in an outline format using predefined structure. The SEA-E, DGE, and our proposed \emph{TreeReview} generate both textual assessments and numerical ratings (Soundness, Presentation, Contribution, and Overall Rating) for papers, while the other methods only generate textual reviews. 

For feedback comments generation, we adopt the following methods: 1) \textbf{D}irect \textbf{P}rompting that identifies paper \textbf{W}eaknesses (\textsc{DPW}) from \citet{lou2024aaar}; 2) Multi-agent collaboration framework MARG \cite{d2024marg}, and its variant without the refinement stage (MARG-\textsc{Base}).

In addition, we include two ablation variants of our \emph{TreeReview} (see §\ref{sec:ablation_study} for details).

\vspace{3pt}

\noindent \textbf{Dataset.} \quad We construct an evaluation benchmark comprising 40 ICLR-2024 papers and 40 NeurIPS-2023 papers along with their corresponding human reviews. For fair comparison, these papers are sampled from the test set of SEA \cite{yu-etal-2024-automated}. To ensure balanced evaluation, we maintain an equal ratio of accepted and rejected papers while maximizing topical diversity across the samples. For the comments generation task, we extract lists of major feedback comments from human reviews, following the procedure of \citet{d2024marg}, to serve as references. More sampling and processing details can be found in Appendix \ref{appendix:benchmark_construction_details}.

\vspace{3pt}

\noindent \textbf{Implementation Details.} \quad For SFT baselines, we utilize the released model weights with their original inference parameters. For other baselines and our \emph{TreeReview}, we employ the \texttt{Gemini-2.0-Flash} (version \texttt{gemini-2.0-flash-001}) via API calls. We set the temperature to 0 and the maximum output length to 32,768 tokens. 

\vspace{3pt}

\noindent\textbf{Hyperparameters Setup.} \quad In this work, the maximum depth of the review question tree ($D_\text{max}$) is set to 4. We employ a depth-aware configuration to control the question decomposition where the maximum number of sub-questions per non-leaf question at depth $l\in \{1,2,\dots,D_\text{max}-1\}$, denoted as $W_\text{max}^l$, follows $W_\text{max}^l = W_\text{max}^{l-1} - 1$ with $W_\text{max}^1 = 5$. This setup is based on the intuition that deeper-level questions become increasingly specific and require fewer sub-questions. During dynamic expansion, a maximum of $W_{\text{max}}^{\raisebox{0.5ex}{\scriptsize{exp}}}=2$ follow-up questions can be generated per intermediate question. For leaf question answering, paper chunks are sized at $L=1024$ tokens, with the top-$k=3$ most relevant chunks selected as context. Implementation details for all baselines are provided in Appendix \ref{appendix:baseline_implementation}.

\subsection{Full Review Generation Task.}
\label{sec:full_review}
\noindent \textbf{Evaluation Setup.} \quad Instead of using conventional text similarity metrics such as ROUGE \cite{lin-2004-rouge} and BERTScore \cite{DBLP:conf/iclr/ZhangKWWA20}, which fail to capture the nuanced qualities of reviews (see analysis in Appendix \ref{appendix:simple_metrics_evaluation}), we adopt the LLM-as-Judge approach, which has demonstrated effectiveness for evaluating complex generation tasks \cite{gao2024reviewer2,li2024generation,wang2024bpo}.

Specifically, we implement a score-based evaluation procedure using \texttt{Gemini-2.5-Pro} (version \texttt{gemini-2.5-pro-exp-0325}) to rate system-generated reviews on a 0-10 scale across eight dimensions: \textit{Comprehensiveness}, \textit{Technical Depth}, \textit{Clarity}, \textit{Constructiveness}, \textit{Specificity}, \textit{Evidence Support}, \textit{Consistency}, and the \textit{Overall Quality}. This approach enables more meaningful and fine-grained quality assessment of reviews. To ensure reliable evaluation, we conduct three independent scoring runs with temperature 0.1 and average the results as final scores. 

Additionally, we conduct a quantitative analysis on paper ratings by calculating the Mean Absolute Error (MAE) and the Mean Squared Error (MSE) between system-assigned and the average ground-truth ratings, which can serve as a measure of the alignment between methods and human reviewers.

For more evaluation settings, including detailed definitions of LLM scoring dimensions, please refer to Appendix \ref{appendix:evaluation_details}. 

\begin{figure}[ht]
    \centering
    \includegraphics[width=\columnwidth]{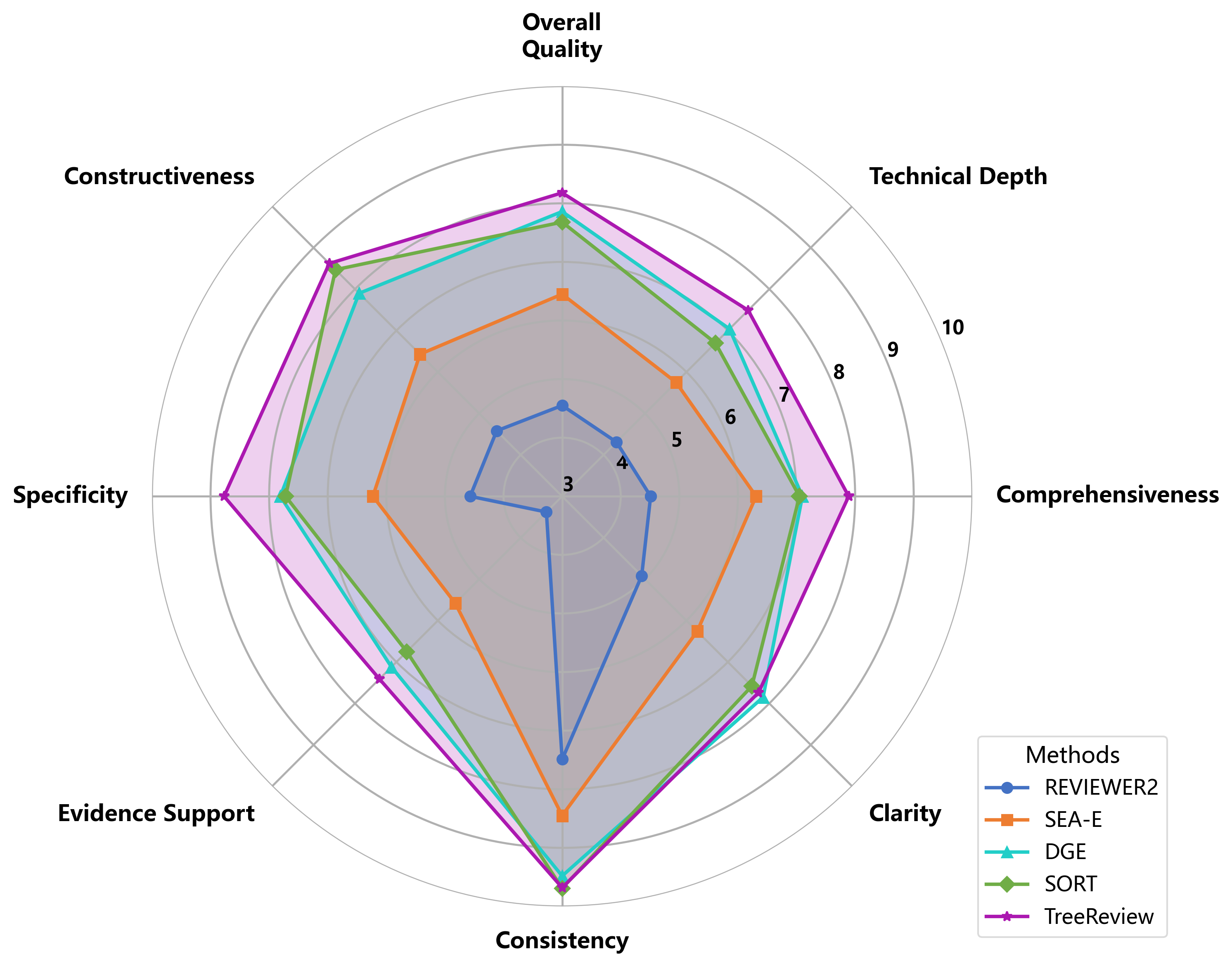}
    \caption{LLM evaluation scores across quality dimensions for all methods.}
    \label{fig:llm_score}
\end{figure}

\vspace{3pt}

\noindent \textbf{Results.} \quad As shown in Fig.~\ref{fig:llm_score}, our \emph{TreeReview} framework achieves the highest overall quality score (8.18) and substantially outperforms all baselines across most quality dimensions, especially in key dimensions such as specificity (\textuparrow 12.27\% over the best baseline), comprehensiveness (\textuparrow 11.22\%), and technical depth (\textuparrow 6.45\%). These gains stem from our divide-and-conquer strategy, which focuses attention on detailed paper content while ensuring coverage through systematic aggregation. 

\begin{table}[ht]
\centering
\setlength\tabcolsep{6.5pt}
\fontsize{8}{7.5}\selectfont

\resizebox{\columnwidth}{!}{
\renewcommand{\arraystretch}{1.3}
\begin{tabular}{l
>{\centering\arraybackslash}p{10pt}
>{\centering\arraybackslash}p{10pt}
>{\centering\arraybackslash}p{10pt}
>{\centering\arraybackslash}p{10pt}
>{\centering\arraybackslash}p{10pt}
>{\centering\arraybackslash}p{10pt}
>{\centering\arraybackslash}p{10pt}
>{\centering\arraybackslash}p{10pt}
}
\toprule[1.2pt]
\multirow{2}{*}{\textbf{Method}} & \multicolumn{4}{c}{MAE} &
\multicolumn{4}{c}{MSE}\\
\cmidrule(lr){2-5} \cmidrule(lr){6-9}
& S. & P. & C. & R. & S. & P. & C. & R. \\
\midrule
DGE & 1.03 & 0.91 & 1.21 & 2.26 & 1.29 & 1.05 & 1.74 & 6.25 \\
SEA-E & \textbf{0.42} & 0.41 & \textbf{0.48} & \textbf{1.17} & \textbf{0.30} & \textbf{0.30} & \textbf{0.37} & \textbf{2.30} \\ 
\midrule
\emph{TreeReview}\textsubscript{$-$\textsc{dec}} & 0.94 & 0.74 & 0.95 & 1.88 & 1.10 & 0.80 & 1.20 & 4.53 \\
\emph{TreeReview}\textsubscript{$-$\textsc{exp}} & 0.55 & \textbf{0.40} & \textbf{0.49} & 1.40 & 0.48 & \textbf{0.30} & \textbf{0.37} & 3.04 \\
\addlinespace[2pt]
\hdashline \addlinespace[2pt]
\emph{TreeReview} & \textbf{0.46} & \textbf{0.36} & \textbf{0.49} & \textbf{1.17} & \textbf{0.35} & \textbf{0.24} & \textbf{0.37} & \textbf{2.12} \\

\bottomrule[1.2pt]
\end{tabular}}
\caption{Results of quantitative analysis on paper ratings. Abbreviations: S.=Soundness, P.=Presentation, C.=Contribution, R.=Overall Rating.}
\label{tab:full_mae_mse}
\end{table}

Among baselines, DGE performs competitively but suffers from focus dilution due to long contexts, resulting in lower comprehensiveness (7.10) and constructiveness (7.90) scores. While SORT excels in constructiveness (8.47), its outline-focused strategy compromises depth and specificity. Fine-tuned models (\textsc{Reviewer2} and SEA-E) consistently underperform across all dimensions, likely due to their limited parameter scale and tendency to mimic surface patterns rather than engaging in critical analysis.

Interestingly, all methods achieve relatively higher scores on consistency than on other dimensions, indicating that maintaining internal coherence is less challenging than providing specific, in-depth feedback. Notably, \emph{TreeReview}'s superior performance in evidence support (\textuparrow 4.16\%) offers practical value by linking claims to specific paper content, facilitating efficient review verification and refinement by human reviewers.

Besides, the intraclass correlation coefficient (ICC) \cite{shrout1979intraclass} across the three scoring runs is 0.9642, indicating strong consistency among LLM judgments.

For paper rating analysis, as shown in Table \ref{tab:full_mae_mse}, \emph{TreeReview} and SEA-E both achieve the lowest level of MAE and MSE across all rating dimensions, demonstrating strong alignment with human reviewer assessments. Notably, while numerical prediction tasks can typically benefit from specialized fine-tuning, \emph{TreeReview} matches or even surpasses (e.g., 2.12 vs. 2.30 MSE for Overall Rating) the performance of the fine-tuned SEA-E. However, the prompt-based DGE method exhibits substantially larger deviations across all rating dimensions, with its MSE reaching 6.25 for Overall Rating.

\subsection{Feedback Comments Generation Task}
\label{sec:feedback_comments_evaluation}
\noindent \textbf{Evaluation Setup.} \quad
We evaluate feedback comments on two key dimensions: \textit{specificity} and \textit{alignment with human reviewer feedback}. We quantify specificity using the ITF-IDF metric introduced by \citet{du-etal-2024-llms}, and a higher ITF-IDF indicates more diverse and unique content in the generated comments. To evaluate alignment, we employ two approaches: 1) Leveraging embedding models to calculate semantic similarity-based metrics \cite{lou2024aaar}, namely SN-Precision, SN-Recall, and SN-F1, and 2) LLM-based alignment evaluation \cite{d2024marg} using \texttt{Gemini-2.5-Pro} to perform many-to-many matching between generated and reference comments. Since reviewers typically provide feedback from different perspectives \cite{yu-etal-2024-automated}, we merge comments from multiple reviewers into an integrated reference set, creating a comprehensive ground truth. Further details are provided in the Appendix \ref{appendix:benchmark_construction_details} and \ref{appendix:evaluation_details}.

\vspace{3pt}

\noindent \textbf{Results.} \quad Results in Table \ref{tab:comments_result} demonstrate \emph{TreeReview}'s superior performance across both specificity and alignment metrics. \emph{TreeReview} achieves the highest precision (32.10\%) in LLM-based alignment, outperforming all baselines. While the strong baseline MARG shows higher recall, \emph{TreeReview} delivers better balance and exceeds MARG by 5.7\% in pseudo-jaccard. Semantic similarity-based alignment evaluation shows consistent results, with \emph{TreeReview} obtaining the highest SN-Precision (47.99\%) and competitive SN-F1 (48.83\%). For specificity, \emph{TreeReview} attains the second-highest ITF-IDF score (4.62), only behind MARG without refinement (5.37\%), which sacrifices alignment for diversity. These results indicate \emph{TreeReview} generates comments that accurately capture human reviewer concerns while maintaining good coverage.

\begin{table*}[t]
\centering
\setlength\tabcolsep{6.5pt}
\fontsize{8}{7.5}\selectfont

\renewcommand{\arraystretch}{1.2}
\begin{tabular}{l
>{\centering\arraybackslash}p{40pt}
>{\centering\arraybackslash}p{40pt}
>{\centering\arraybackslash}p{40pt}
>{\centering\arraybackslash}p{40pt}
>{\centering\arraybackslash}p{40pt}
>{\centering\arraybackslash}p{40pt}
>{\centering\arraybackslash}p{40pt}}
\toprule[1.2pt]
\multicolumn{1}{c}{\multirow{2}{*}{\textbf{Method}}} & \multicolumn{3}{c}{LLM-based alignment} & \multicolumn{3}{c}{Semantic similarity} & \multicolumn{1}{c}{\multirow{2}{*}{ITF-IDF}}\\
\cmidrule(lr){2-4} \cmidrule(lr){5-7} 
& Precision & Recall & Jaccard & SN-P & SN-R & SN-F1\\
\midrule

DPW & 9.47 & 9.87 & 5.31 & 43.72 & \textbf{53.59} & 48.05 & \textbf{4.48} \\
SORT & 22.66 & 10.75 & 8.17 & \textbf{45.70} & 47.21 & 46.30 & 3.45 \\
MARG-\textsc{Base} & 6.02 & 15.40 & 4.53 & 36.37 & 51.37 & 42.43 & 5.37 \\
MARG & 13.38 & \textbf{23.98} & 9.63 & 45.13 & \textbf{55.01} & \textbf{49.42} & 4.22 \\

\midrule
\emph{TreeReview}\textsubscript{$-$\textsc{dec}} & 13.49 & 15.76 & 7.93 & 43.92 & 52.68 & 47.71 & 3.76\\
\emph{TreeReview}\textsubscript{$-$\textsc{exp}} & \textbf{26.06} & 19.58 & \textbf{12.98} & 44.30 & 50.28 & 46.93 & 4.01\\[1ex]
\hdashline
\emph{TreeReview} & \textbf{32.10} & \textbf{21.68} & \textbf{15.33} & \textbf{47.99} & 50.32 & \textbf{48.83} & \textbf{4.62} \rule{0pt}{3ex} \\
\bottomrule[1.2pt]
\end{tabular}
\caption{Performance comparison of different methods on feedback comment generation across specificity and alignment metrics.}
\label{tab:comments_result}
\end{table*}

\begin{figure}[ht]
    \centering
    \includegraphics[width=\columnwidth]{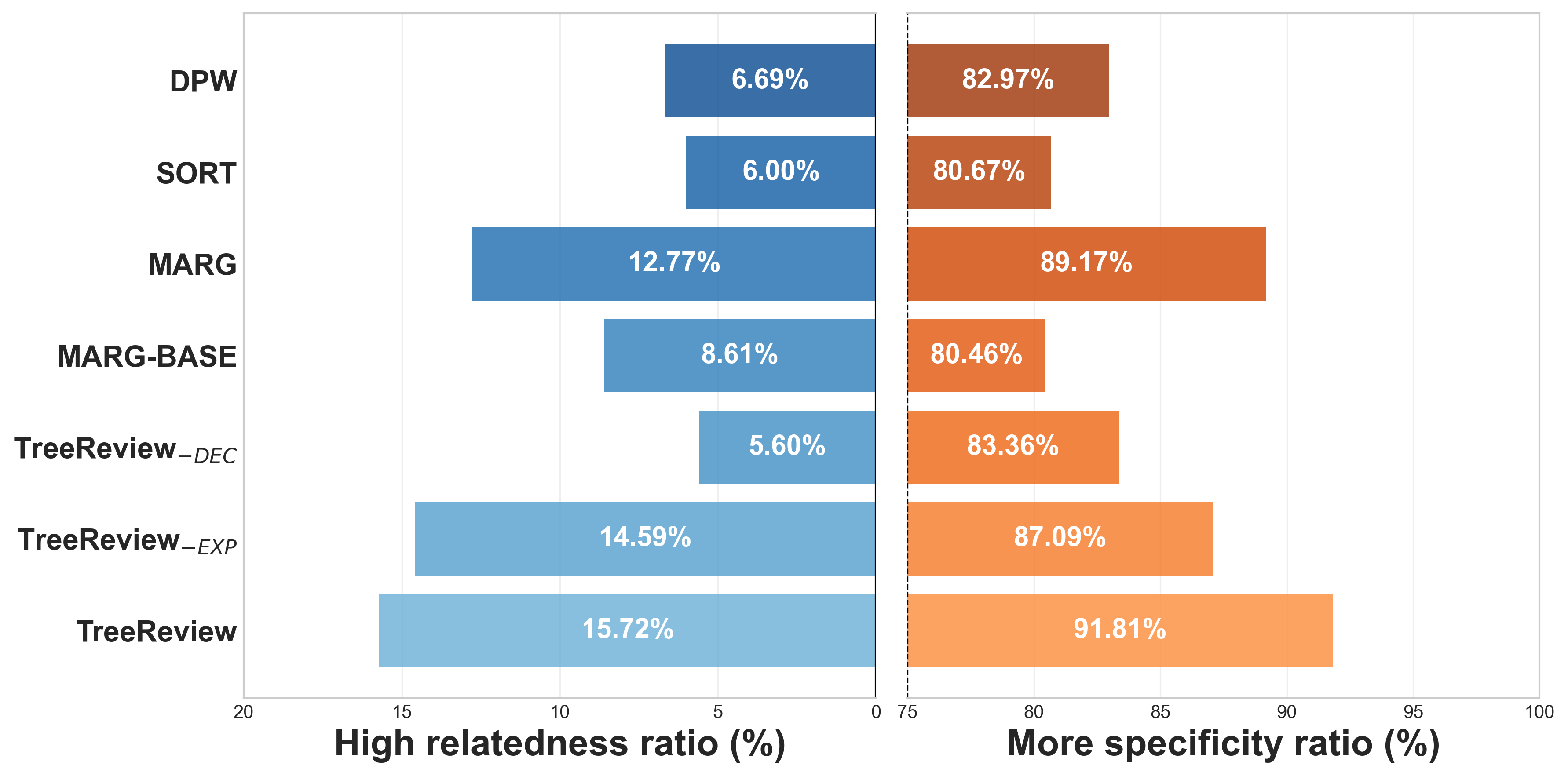}
    \caption{Proportion of generated comments judged as ``highly related'' and ``more specific'' in LLM-based alignment evaluation across different methods. }
    \label{fig:llm_align_ratio}
\end{figure}

In Fig.~\ref{fig:llm_align_ratio}, we further report the proportion of aligned comments judged as ``highly related'' and ``more specific'' in the LLM-based alignment evaluation. \emph{TreeReview} achieves the highest proportion of ``highly related'' comments (15.72\%) among all methods. Consistent with ITF-IDF results, \emph{TreeReview} yields the highest ``more specific'' ratio, suggesting that \emph{TreeReview} can produce more paper-specific and informative feedback.

\subsection{Human Evaluation}
To complement our automatic evaluation, we conduct a human evaluation with 20 papers randomly sampled from the test set. Five evaluators with experience reviewing for top NLP/ML conferences are recruited to evaluate pairs of reviews and sets of feedback comments generated by different methods. Each pair is assessed by two different evaluators to ensure reliability, with the evaluation procedure and criteria detailed in the Appendix \ref{appendix:human_eval_details}.

\begin{figure}[ht]
    \centering
    \includegraphics[width=\columnwidth]{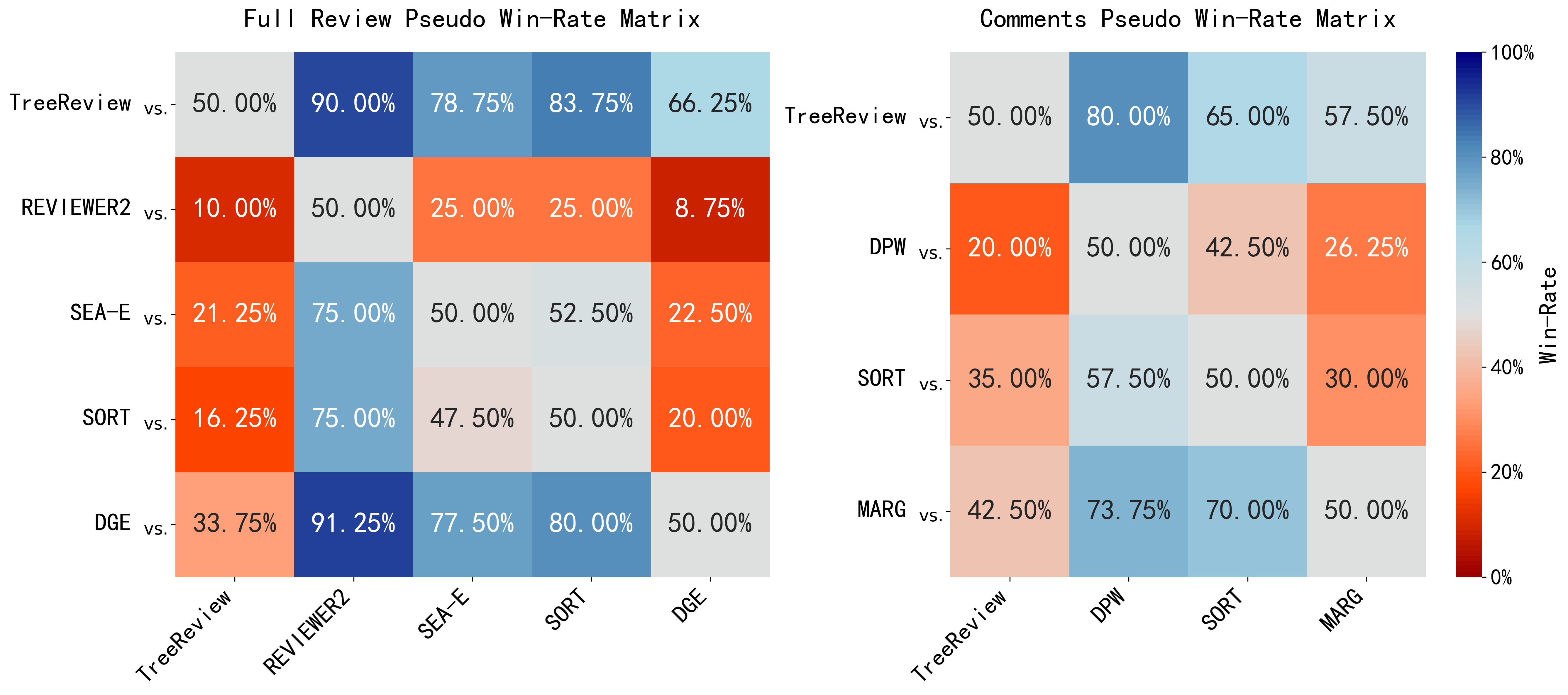}
    \caption{Human evaluation pair-wise win-rates for full review and feedback comments generation tasks.}
    \label{fig:win_rate}
\end{figure}

Fig.~\ref{fig:win_rate} shows \emph{TreeReview} consistently outperforming baselines on human evaluation across two review tasks. For full reviews, \emph{TreeReview} achieves win-rates between 66.25\% (against DGE) and 90.00\% (against \textsc{Reviewer2}). For feedback comments, \emph{TreeReview} surpasses the strong MARG baseline (57.50\% vs. 42.50\%). The high inter-evaluator agreement (overall agreement of 0.75 and Cohen $\kappa$ of 0.70) indicates reliable human judgments. In addition, we have calculated the Spearman's rank correlation coefficient between the "Overall Quality" scores assigned by the LLM and the aggregated win-rate scores derived from human pairwise comparisons. The results show strong consistency ($\rho \approx 0.90$, $p < 0.05$), further supporting the validity of LLM-based assessment. These results demonstrate that our framework generates reviews and comments that better align with expert preferences compared to baseline methods.

\subsection{Domain Generalization Analysis}

\begin{table*}[htbp]
\centering
\setlength\tabcolsep{6.5pt}
\fontsize{8}{7.5}\selectfont

\begin{tabular}{lccccccccc}
\toprule[1.2pt]
\textbf{Method} & Comp. & Tech. & Clari. & Const. & Spec. & Evi. & Cons. & Overall Quality \\
\midrule
DGE & 6.73 & 6.11 & 7.43 & 7.63 & 7.58 & 6.48 & 9.38 & 7.26 \\
SORT & 7.06 & 6.39 & 7.72 & 8.64 & 7.74 & 6.30 & \textbf{9.70} & 7.63 \\
\textbf{\emph{TreeReview}} & \textbf{7.64} & \textbf{7.83} & \textbf{8.38} & \textbf{8.80} & \textbf{8.62} & \textbf{7.85} & 9.58 & \textbf{8.31} \\
\bottomrule[1.2pt]
\end{tabular}
\caption{LLM evaluation scores of \emph{TreeReview} and representative baseline methods across quality dimensions on multi-domain dataset. Abbreviations: Comp. = Comprehensiveness, Tech. = Technical Depth, Clari. = Clarity, Const. = Constructiveness, Spec. = Specificity, Evi. = Evidence Support, Cons. = Consistency.}
\label{tab:full_review_nature}
\end{table*}


\begin{table*}[htbp]
    \centering
    \setlength\tabcolsep{6.5pt}
    \fontsize{8}{7.5}\selectfont
    
    \begin{tabular}{lcccccccc}
        \toprule[1.2pt]
        
        \multirow{2}{*}{\textbf{Method}} & \multicolumn{5}{c}{LLM-based alignment} & \multicolumn{3}{c}{Semantic similarity} \\
       
        \cmidrule(lr){2-6} \cmidrule(lr){7-9}
        & Precision & Recall & Jaccard & High Relat. \% & More Spec. \% & SN-P & SN-R & SN-F1 \\
        \midrule
        DPW & 12.85 & 27.73 & 9.68 & 15.87 & 87.07 & 40.83 & \textbf{58.37} & 47.77 \\
        SORT & 27.92 & 16.71 & 11.71 & 22.22 & 82.10 & 47.02 & 50.35 & 48.34 \\
        \textbf{\emph{TreeReview}} & \textbf{34.62} & \textbf{29.11} & \textbf{17.68} & \textbf{25.85} & \textbf{93.72} & \textbf{52.26} & 56.93 & \textbf{54.20} \\
        \bottomrule[1.2pt]
    \end{tabular}
    \caption{LLM-based alignment and semantic similarity evaluation against ground-truth comments for actionable feedback comments generation on multi-domain dataset. Abbreviations: High Relat.= High Relatedness, More Spec.= More Specificity.}
    \label{tab:actionable_feedback_nature}
\end{table*}

While our benchmark primarily focuses on the AI/ML domain, we extended our evaluation to papers from other scientific fields to assess the robustness and generalization of our approach. Specifically, we curate 40 papers from \textit{Nature Communications}, covering  biology (13), climate science (14), and quantum science (13). We employ the same evaluation metrics and representative baselines as described in §\ref{sec:full_review} and §\ref{sec:feedback_comments_evaluation}, thereby maintaining a consistent evaluation framework.

Tables \ref{tab:full_review_nature} and \ref{tab:actionable_feedback_nature} present the experimental results on this multi-domain dataset. For full review generation, \emph{TreeReview} achieves the highest overall quality score of 8.31, outperforming SORT (7.63) and DGE (7.26), with notable improvements in technical depth (7.83 vs. 6.39) and evidence support (7.85 vs. 6.30). For actionable feedback generation, \emph{TreeReview} maintains superior performance with the highest precision (34.62\%)and Jaccard similarity (17.68\%). 

The performance observed on this  dataset closely align with our previous findings, with \emph{TreeReview} consistently outperforming baselines across almost all evaluation metrics. This consistency validates that our method's advantages stem from fundamental improvements in review generation rather than domain-specific optimizations, supporting the broader applicability of hierarchical question decomposition for automated peer review across scientific disciplines.

\subsection{Ablation Study}
\label{sec:ablation_study}
To evaluate the contributions of key components in \emph{TreeReview}, we conduct ablation experiments on two variants: 1) \emph{TreeReview}\textsubscript{$-$\textsc{dec}}, which removes the question tree decomposition and answer aggregation, reducing the framework to direct prompting of the LLM for the review tasks; 2) \emph{TreeReview}\textsubscript{$-$\textsc{exp}}, which removes the dynamic question expansion mechanism, restricting the framework to the initial question tree.

Results in Tables~\ref{tab:full_mae_mse} and~\ref{tab:comments_result} show that \emph{TreeReview}\textsubscript{$-$\textsc{dec}} significantly degrades both quantitative accuracy and qualitative review quality, highlighting the critical role of the divide-and-conquer reasoning.
For \emph{TreeReview}\textsubscript{$-$\textsc{exp}}, rating prediction remains relatively robust (with minor increases in MAE/MSE), but the quality of feedback comments drops notably, with reduced ability to identify critical issues (Jaccard score decreasing from 15.33\% to 12.98\%) and less specific comments (ITF-IDF dropping from 4.62 to 4.01). Statistical analysis reveals that the dynamic expansion mechanism triggers expansion for 38.54\% of non-leaf questions on average, generating 25.6 additional questions per review, enabling deeper probing of ambiguous or underexplored areas.
Further insights from our case study (Appendix~\ref{appendix:case_study}) demonstrate that many highly aligned and specific comments stem directly from fine-grained, dynamically expanded questions. 

These findings collectively underscore the key role of both hierarchical decomposition and dynamic expansion in \emph{TreeReview} for providing comprehensive, specific, and expert-aligned feedback.

\begin{table}
\centering
\setlength\tabcolsep{6.5pt}
\fontsize{8}{7.5}\selectfont

\resizebox{\columnwidth}{!}{
\renewcommand{\arraystretch}{1.3}
\begin{tabular}{l
>{\centering\arraybackslash}p{50pt}
>{\centering\arraybackslash}p{50pt}
>{\centering\arraybackslash}p{50pt}
}
\toprule[1.2pt]
\multirow{2}{*}{\textbf{Method}} & Input tokens/paper & Output tokens/paper & Total tokens/paper \\
\midrule
MARG & 2,192,910 & 121,141 & 2,314,052 \\
MARG-\textsc{Base} & 963,027 & 44,581 & 1,007,608\\
\emph{TreeReview} & 419,929 & 39,039 & 458,968 \\
\bottomrule[1.2pt]
\end{tabular}
}
\caption{Statistics of per-paper average token usage.}
\label{tab:cost_analysis}
\end{table}

\subsection{Cost Analysis}
In this section, we compare the computational efficiency of MARG, MARG-\textsc{Base}, and our proposed method on the feedback comments generation task. As shown in Table \ref{tab:cost_analysis}, \emph{TreeReview} substantially reduces the per-paper average token usage, with a decrease of 80.2\% compared to MARG and 54.4\% compared to MARG-\textsc{Base}. Despite this, as demonstrated in §\ref{sec:feedback_comments_evaluation}, our method still maintains superior or competitive performance across evaluation metrics. These efficiency advantages translate to shorter processing times and lower API costs, making \emph{TreeReview} more practical for assisting the real-world review process.

\section*{Conclusion}
In this paper, we introduce \emph{TreeReview}, a novel framework designed to address key challenges of LLM-based paper review through a dynamic, hierarchical question-answering architecture. The extensive experiments on our constructed benchmark demonstrate that \emph{TreeReview} shows superiority in providing in-depth and helpful review feedback compared to baselines while maintaining efficiency. Our ablation studies highlight the importance of both the hierarchical decomposition strategy and the dynamic expansion mechanism. \emph{TreeReview} offers a new approach to leveraging LLMs in assisting the peer review process and also potentially benefits more tasks involving deep comprehension of long text.

\section*{Limitations}
Despite the promising results of \emph{TreeReview}, several limitations remain to be addressed in future work:

\noindent \textbf{LLM Limitations and Risks.} \quad \emph{TreeReview} relies on LLMs for both question generation and answer synthesis. The overall review quality is thus bounded by the backend LLMs’ capabilities, such as uneven knowledge coverage of highly specialized domains. Furthermore, LLMs are prone to hallucination, which could result in propagation of factual incorrect statements in some cases. Although the evidence-based answering aggregation helps mitigate this issue by grounding answers in the paper's content, it cannot entirely eliminate the risk of producing misleading feedback.

\noindent \textbf{Multimodal Input Consideration.} \quad In this work, we do not incorporate figures as model input, as we believe that their corresponding captions and analysis within the paper already provide the essential information needed for \emph{TreeReview} to generate high-quality feedback. However, given the rapid advancement of multimodal models, their potential merits deserve attention, and we plan to evaluate in future work whether incorporating such models would provide substantial benefits that outweigh their computational costs.

\section*{Ethical Considerations}
While \emph{TreeReview} demonstrates promising capabilities in generating high-quality scientific peer reviews, we emphasize that it is designed to assist rather than replace human reviewers. Its primary intention is to aid authors in refining manuscripts before submission and to provide supplementary insights for reviewers facing heavy workloads. However, automatic review generation introduces ethical risks, most notably, the potential misuse of generated reviews as substitutes for genuine expert assessment in formal reviewing workflows. Such misuse could undermine fairness, transparency, and trust in peer review. To mitigate these concerns, we strongly discourage deploying \emph{TreeReview} outputs as official, standalone reviews or final recommendations. Instead, all automatically generated feedback should remain subject to human interpretation and oversight. Additionally, the datasets used in this work are publicly available and are intended solely for legitimate research purposes. 

\section*{Acknowledgements}
This work was supported by the Young Backbone Talent Program of the National Science Library, Chinese Academy of Sciences (No. E5290902) and Theme-based Research Scheme (TRS) project T45-701/22-R. We would like to thank all the anonymous reviewers for their valuable comments and constructive feedback.

\bibliography{custom}


\appendix

\section{More details of \emph{TreeReview}}
\label{appendix:method_detail}
\subsection{Question-aware Chunk Reranking}
\label{appendix:chunk_rerank_details}
As described in §\ref{sec:bottom_up_aggregation}, answering leaf questions ($q_i^\text{leaf}$) requires identifying the most relevant content segments (chunks) from the paper $\mathcal{P}$ to serve as focused context for the Answer Synthesizer agent $M_a$. This appendix provides implementation details for the question-aware chunk reranking process used to select the top-$k$ relevant chunks.

\vspace{3pt}
\noindent\textbf{Paper Chunking} \quad
We first segment the full paper $\mathcal{P}$ into chunks. We set the target chunk size to $L=1024$ tokens with token counts measured using \textit{tiktoken}\footnote{\url{https://github.com/openai/tiktoken}}. Chunks are allowed to exceed this size to avoid truncating paragraphs mid-content. This approach ensures the semantic coherence of each chunk while maintaining reasonable context windows. To enhance contextual awareness and support evidence citation, we prepend section hierarchy information to each chunk in the format of ``Section Title > Sub-Section Title > \dots''. This provides the LLM with relative positional cues of the chunk within the paper structure and facilitates precise referencing of content during answer generation.

\vspace{3pt}
\noindent\textbf{Chunk Reranking} \quad
To identify the most relevant chunks for a given leaf question $q_i^\text{leaf}$, we adopt the question-aware context compression technique presented in LongLLMLingua \cite{jiang-etal-2024-longllmlingua}. Specifically, we evaluate the relevance of each chunk $\text{chunk}_j$ by computing the perplexity $ppl_j$ of the question $q_i^\text{leaf}$ conditioned on $\text{chunk}_j$, where higher perplexity means less relevance:
\begin{equation}
ppl_j = - \log{p(q_i, x^\text{restrict} \mid \text{chunk}_j)}
\end{equation}

The restrictive statement ``We can get the answer to this question in the given documents'' serves as a regularization term to mitigate hallucination and strengthen the connection between the question and context.
The chunks are ranked according to the calculated perplexity, and the top-$k$ chunks (we set $k=3$ in this work) with the lowest perplexity are selected to form the context provided to the Answer Synthesizer $M_a$ for answering the leaf question $q_i^\text{leaf}$.

\vspace{3pt}

\noindent\textbf{Implementation} \quad
For the language model used to compute perplexity, we employ Llama-3.1-8B-Instruct\footnote{\url{https://huggingface.co/meta-llama/Llama-3.1-8B-Instruct}}, which provides a good balance between performance and efficiency. The inference is performed using two NVIDIA RTX 4090 GPUs.

\subsection{Intermediate Question Processing}
The procedure for resolving intermediate questions encompasses both answer synthesis and dynamic question expansion (§\ref{sec:bottom_up_aggregation}).
For each intermediate question $q_i^\text{inter}$ (i.e., non-leaf and non-root), the Answer Synthesizer $M_a$ receives the question itself and all its current sub-question-answer pairs ${(q_{i,j}, a_{i,j})}_{j=1}^{\bar{n}_i}$.  The core task is to determine if the collective evidence provided by these sub-answers is sufficient to comprehensively address $q_i^\text{inter}$. We employ Chain-of-Thought (CoT) prompting to guide $M_a$ through this decision process and the subsequent action. Specifically, we explicitly instruct $M_a$ to output its reasoning steps before generating the final output. When $M_a$ determines that the available information is sufficient, it synthesizes an answer by integrating and abstracting insights from the sub-answers. Otherwise, $M_a$ generates follow-up questions targeting the gaps. The detailed prompts are provided in the Appendix \ref{appendix:treereview_prompts}.

\section {Experimental Setting Details}
\label{appendix:experimental_details}
\subsection {Benchmark Construction}
\label{appendix:benchmark_construction_details}
To construct a robust evaluation benchmark for our experiments, we sample 40 ICLR-2024 papers and 40 NeurIPS-2023 papers, along with their corresponding human-written reviews, from the test set of the SEA dataset \cite{yu-etal-2024-automated}.\footnote{\url{https://huggingface.co/datasets/ECNU-SEA/SEA_data}, licensed by the Apache License 2.0.} For the full paper content, we use the pre-processed Markdown files provided by the SEA, which are converted from the paper PDF and retain text, tables, and equations while excluding visual elements like figures. 

\begin{table*}[t]
\centering
\setlength\tabcolsep{6.5pt}
\fontsize{8}{7.5}\selectfont

\renewcommand{\arraystretch}{1.2}
\begin{tabular}{l
>{\centering\arraybackslash}p{50pt}
>{\centering\arraybackslash}p{50pt}
>{\centering\arraybackslash}p{50pt}
>{\centering\arraybackslash}p{50pt}
>{\centering\arraybackslash}p{40pt}}
\toprule[1.2pt]
\multicolumn{1}{c}{} & \multicolumn{2}{c}{NeurIPS-2023} & \multicolumn{2}{c}{ICLR-2024} & \multicolumn{1}{c}{Total} \\
\cmidrule(lr){2-3} \cmidrule(lr){4-5}
& Full & \cellcolor{gray!10}Human Eval. & Full & \cellcolor{gray!10}Human Eval. & \\
\midrule
\# papers & 40 & \cellcolor{gray!10}10 & 40 & \cellcolor{gray!10}10 & 80 \\
\% accepted & 50.0\% & \cellcolor{gray!10}50.0\% & 50.0\% & \cellcolor{gray!10}50.0\% & 50.0\% \\
\# tokens per paper & 16,351 & \cellcolor{gray!10}16,275 & 21,909 & \cellcolor{gray!10}22,326 & 19,130 \\
\# reviews per paper & 4.5 & \cellcolor{gray!10}4.4 & 3.9 & \cellcolor{gray!10}4.2 & 4.2 \\
\# tokens per review & 698 & \cellcolor{gray!10}733 & 664 & \cellcolor{gray!10}645 & 682 \\
\# comments per review & 3.4 & \cellcolor{gray!10}3.1 & 3.8 & \cellcolor{gray!10}3.4 & 3.6 \\
\# tokens per comment & 44 & \cellcolor{gray!10}49 & 43 & \cellcolor{gray!10}43 & 43 \\
\# merged comments per paper & 9.3 & \cellcolor{gray!10}9.8 & 9.8 & \cellcolor{gray!10}9.0 & 9.5 \\
\# tokens per merged comment & 69 & \cellcolor{gray!10}64 & 62 & \cellcolor{gray!10}65 & 65 \\
\bottomrule[1.2pt]
\end{tabular}
\caption{Statistical overview of our evaluation benchmark, covering both NeurIPS-2023 and ICLR-2024 venues. Full: the main experimental set; \colorbox{gray!10}{Human Eval.: subset for human evaluation}; Total: overall statistics across all papers.}
\label{tab:dataset_statistics}
\end{table*}

We employ a stratified sampling approach to ensure balanced distribution across acceptance decisions, selecting 20 accepted and 20 rejected papers from each venue. To maximize the diversity of topics within our dataset, we implement a diversity-aware sampling strategy based on the Min-Max algorithm:
\begin{itemize}[leftmargin=*]
    \item[-]We first randomly select an initial paper from each venue-decision category.
    \item[-]We utilize the \texttt{multilingual-e5-small}\footnote{\url{https://huggingface.co/intfloat/multilingual-e5-small}} embedding model to compute semantic representations of papers based on the concatenation of paper title and abstract.
    \item[-]For subsequent selections, we identify papers that had no keyword overlap with the already selected papers and maximize the minimum cosine distance of embeddings between the current paper and all previously selected papers. 
\end{itemize}

\begin{figure}[ht]
    \centering
    \includegraphics[width=\columnwidth]{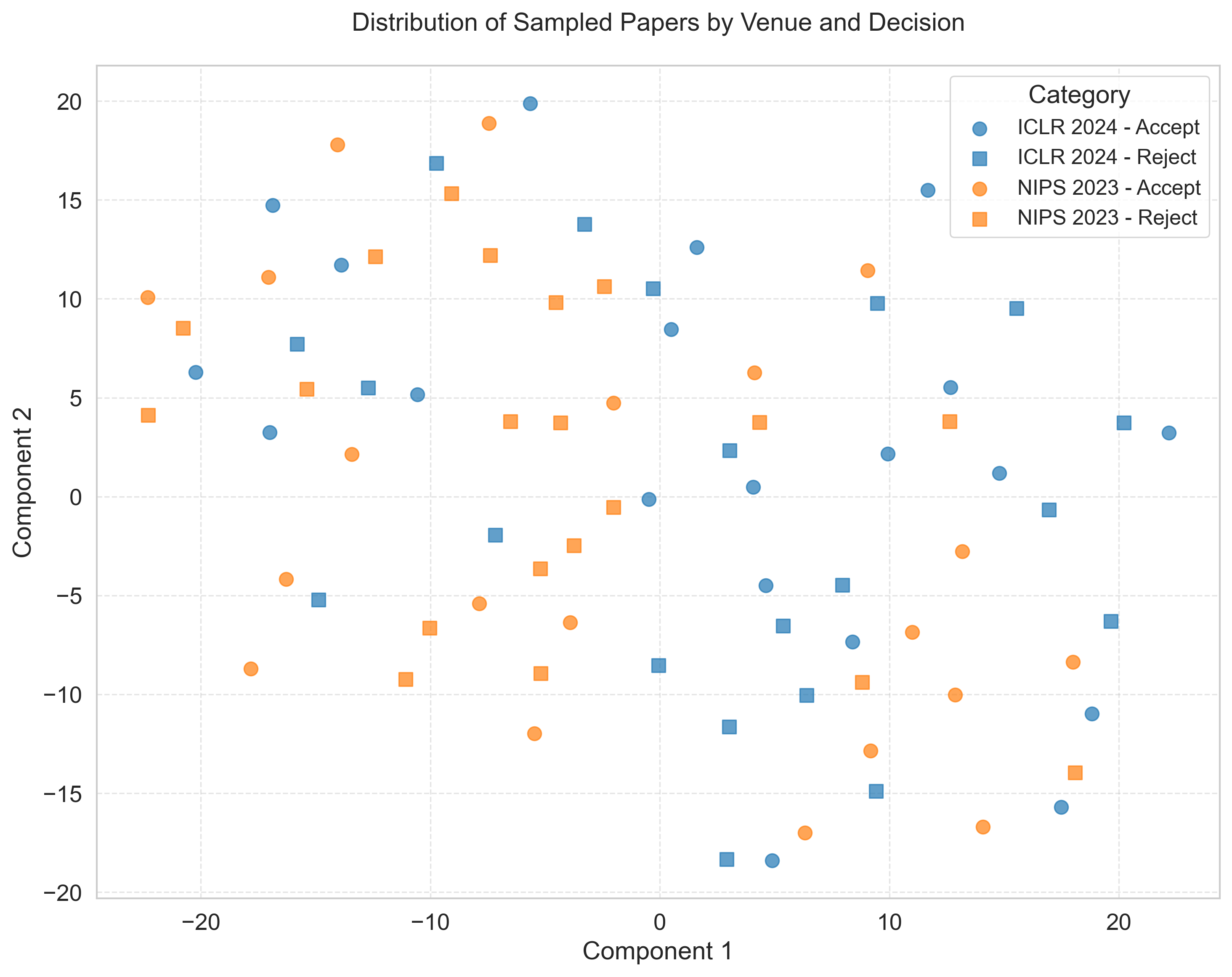}
    \caption{t-SNE visualization of sampled papers showing the diversity of topics across venues and acceptance decisions in our benchmark.}
    \label{fig:paper_diversity}
\end{figure}

To further illustrate the topical diversity of our benchmark, we project the concatenation of the title and abstract of each sampled paper into a high-dimensional semantic space using \texttt{multilingual-e5-small} embedding model and visualize the distribution using the t-SNE technique. As shown in Fig. \ref{fig:paper_diversity}, the sampled papers are evenly distributed across the semantic space, reflecting a broad range of topics. This diversity holds for both accepted and rejected papers across the ICLR-2024 and NeurIPS-2023 venues.

For the full review generation task, we directly use the original human reviews as references, including both textual comments and numerical ratings (Soundness, Presentation, Contribution, and Overall Rating). 

\begin{figure*}
    \centering
    \small
    \begin{tcolorbox}[colframe=black, colback=white, width=\textwidth, boxrule=0.2mm]
\textbf{Instructions}: \\
A user will give you a scientific paper review, and you must make the list of comments made by the reviewer.  Write each specific suggestion or critique that the reviewer makes.  Each item in the list should stand alone as a complete comment, so you may need to paraphrase or adjust comments in order to add context and improve clarity.  However, you should try to preserve the original wording when possible.  Do not reframe comments as reported speech or add attributions.  In addition, you should merge similar comments as needed to ensure that each final comment in your list stands on its own as a fully-contextualized comment.  For example, a reviewer might give a high-level comment like "Experiments are not convincing" and then elaborate on that comment later with a more detailed explanation of how the experiments are unconvincing; in this case, you should merge the two comments into a single comment with all the details.
\\ \hspace*{\fill} \\
Your output should be a JSON object like `{"major": List[str], "minor": List[str]}` where the lists of strings are the lists of review comments.  The "major" comments should be the most important ones, typically regarding the impact and novelty of the work, the correctness of main claims, or anything else that the reviewer suggests is an important factor in accepting the work.  The "minor" comments should be the ones that are just about small details that aren't crucial for the work, such as style and grammar, minor clarifications, or other things that the reviewer indicates aren't important.
\\ \hspace*{\fill} \\
Example: <EXAMPLE>
    \end{tcolorbox}
    \caption{Instructions for extracting feedback comments from human reviews.}
    \label{fig:comments_extraction_prompt}
\end{figure*}

\begin{figure*}
    \centering
    \small
    \begin{tcolorbox}[colframe=black, colback=white, width=\textwidth, boxrule=0.2mm]
\textbf{Instructions}:\\
You will receive review comments from multiple reviewers for the same scientific paper. Each reviewer's feedback is structured as a list of important critiques or suggestions about the paper.

Your task is to merge these multiple sets of feedback comments into a single consolidated list that comprehensively represents all the important feedback. When merging, follow these guidelines:
\begin{enumerate}[leftmargin=*, topsep=0pt, label={\arabic*}.]
\setlength{\itemsep}{0pt}
\setlength{\parsep}{0pt}
\setlength{\parskip}{0pt}
    \item If multiple reviewers mention the same issue, combine them into a single comment that preserves all details, ensuring no duplicate comments. If one reviewer provides a more detailed explanation than another on the same point, include the more comprehensive version with all specifics. For example, a reviewer might give a high-level comment like "Experiments are not convincing" and then elaborate on that comment later with a more detailed explanation of how the experiments are unconvincing; in this case, you should merge the two comments into a single comment with all the details. For example, one reviewer might give a high-level comment like "Experiments are not convincing" while another reviewer raises a similar concern but provides a more detailed explanation like "The experimental setup lacks statistical significance tests and has insufficient sample size." In this case, you should merge these comments into a single comprehensive comment that captures both the general concern and the specific details.
    \item If conflicting comments exist between reviewers, preserve all conflicting viewpoints in the final list, do not attempt to resolve contradictions.
    \item Try to preserve original wording, voice, and phrasing whenever possible, with minimal rewording only when necessary for clarity or to properly merge similar comments. Do not reframe comments as reported speech or add attributions.
    \item Ensure each comment in your final list is fully contextualized and can stand alone as a complete comment.
    \item Do not add new critiques or suggestions that weren't present in the original review comments.
\end{enumerate}
Your output should be a single JSON array of strings, where each string is a complete, consolidated comment. Do not include any numbering, bullet points, or other special markers in the output. The format should be:
\begin{lstlisting}[basicstyle = \ttfamily]
[
  "First consolidated comment",
  "Second consolidated comment",
  "Third consolidated comment",
  ...
]
\end{lstlisting}

    \end{tcolorbox}
    \caption{Instructions for merging multiple sets of feedback comments from different human reviewers.}
    \label{fig:merge_multiple_reviewers_comments}
\end{figure*}

\label{appendix:merge_comments}
For the actionable feedback comments task, we extract lists of major feedback points from human reviews following the procedure described in MARG \cite{d2024marg}, and the instructions are shown in Fig. \ref{fig:comments_extraction_prompt}. Crucially, for the LLM-based alignment evaluation, we differ from the method in MARG, which aligns generated comments against each reviewer's comments individually. Instead, we merge the extracted comments from all reviewers of the same paper into a single, consolidated reference set, utilizing the instructions shown in Fig. \ref{fig:merge_multiple_reviewers_comments}. This merging process combines similar comments while preserving unique perspectives, resulting in a more comprehensive ground truth that captures the full spectrum of expert opinions on each paper. Both the extraction and merging processes are implemented using \texttt{Gemini-2.5-Pro}. We also conduct manual checks on 15 cases and find that \texttt{Gemini-2.5-Pro} reliably extracts nearly all salient insights from the human reviews and accurately merges similar points.

Table~\ref{tab:dataset_statistics} presents detailed statistics of the constructed benchmark, providing an overview of paper, review, and comment distributions across venues and evaluation settings. The substantial average paper length (~20K tokens) presents a significant challenge for LLMs to accurately capture and reason over nuanced paper details while maintaining comprehensive understanding.

\subsection{Baselines Implementation Details}
\label{appendix:baseline_implementation}
\noindent\textbf{\textsc{Reviewer2}} \quad \textsc{Reviewer2} is a two-stage review generation framework designed to enhance the coverage and specificity of generated reviews. It consists of two fine-tuned LLMs: the first model $M_p$ generates aspect prompt based on the paper, and the second model $M_r$ produces the final review based on the paper and the aspect prompt. To facilitate training, \textsc{Reviewer2} introduced a Prompt Generation with Evaluation (PGE) pipeline to annotate existing review datasets with corresponding aspect prompts.

\vspace{3pt}

\noindent\textbf{SEA-E} \quad SEA-E is the Evaluation module within the SEA framework, designed for automated scientific review generation. The key idea behind SEA-E is to fine-tune the LLMs using high-quality, standardized review data rather than potentially biased or partial individual reviews. To achieve this, the SEA framework first utilizes its Standardization module (SEA-S) to integrate multiple raw human reviews for each paper into a single, unified, and comprehensive format, leveraging GPT-4 distillation. SEA-E is then implemented by fine-tuning on the standardized review dataset. 

\vspace{3pt}

\noindent For both \textsc{Reviewer2} and SEA-E, we utilize their released model weights and run on 2 NVIDIA RTX 4090 GPUs. All inference parameters, such as temperature, are configured following the original settings provided in their released codes.

\vspace{3pt}

\noindent\textbf{DGE} \quad This baseline implements a prompt-based approach combining step-by-step review guidelines from top-tier conferences with few-shot examples of human-written reviews, following the methodology of \citet{du-etal-2024-llms} and \citet{lu2024ai}. This method leverages the in-context learning capabilities of LLMs. Specifically, we craft a prompt comprising the ICLR 2024 Reviewer Guide\footnote{\url{ https://iclr.cc/Conferences/2024/ReviewerGuide}} and two exemplar reviews that lean to accept and reject, respectively, as well as the detailed review format. This strategy emulates how human reviewers rely on guidelines and expert examples to formulate their critiques.

\vspace{3pt}

\noindent\textbf{SORT} \quad 
This baseline implements the approach from \citet{liang2024can}, which utilizes predefined section templates to guide the LLM in generating reviews that cover various aspects such as significance and novelty, potential reasons for acceptance and rejection, and suggestions for improvement. The method prompts the LLM to produce reviews in outline format by following these templates, ensuring comprehensive coverage. Furthermore, as part of their evaluation protocol, they also extract comments that focus on potential reasons for rejection from the generated reviews, enabling this approach to serve as a baseline for both full review generation and actionable feedback comments generation tasks within our experimental setup. For implementation, we use the original prompt templates for both review generation and comments extraction.

\vspace{3pt}

\noindent\textbf{MARG} \quad The MARG (Multi-Agent Review Generation) method is a multi-agent framework designed to generate peer-review feedback by leveraging the collaboration of multiple LLM instances. It employs a distributed architecture with a leader agent coordinating tasks, worker agents handling portions of the paper text, and expert agents specializing in sub-tasks to assist the leader agent. MARG also utilizes independent multi-agent groups for different aspects of the review, such as experiments, clarity, and impact. 
However, the original MARG implementation faced challenges with communication errors, such as misplaced SEND MESSAGE markers and excessive use of SEND FULL MESSAGE, leading to inefficient message broadcasting and potential miscommunication among agents. In our implementation, we refine the communication protocol by removing SEND FULL MESSAGE, restricting agents to a single SEND MESSAGE per output, and instructing the agents to broadcast messages only after planning is complete. 
We also include a variant of MARG without the refinement stage to serve as an additional baseline in our experiments.

\section{Evaluation Details}
\label{appendix:evaluation_details}
\subsection{Text Similarity-based Evaluation}
\label{appendix:simple_metrics_evaluation}
\noindent\textbf{Metrics and Setup} \quad
In our preliminary experiments for the full review generation task, we employ two widely-used text similarity metrics, ROUGE (including R-1, R-2, R-L) and BERTScore, to evaluate the quality of generated reviews against human-written reference reviews. We calculate the maximum score across multiple reference reviews for each generated review to account for the diversity of human perspectives. Additionally, inspired by the specificity metric (SPE) from \citet{gao2024reviewer2}, we report the average drop in BERTScore (Avg-Drop) when pairing generated reviews with reference reviews of a different paper, approximated via Monte Carlo sampling over 10 iterations.

\begin{table*}
\centering
\small
\renewcommand{\arraystretch}{1.3}
\begin{tabular}{l
>{\centering\arraybackslash}p{25pt}
>{\centering\arraybackslash}p{25pt}
>{\centering\arraybackslash}p{25pt}
>{\centering\arraybackslash}p{25pt}
>{\centering\arraybackslash}p{25pt}
}
\toprule[1.2pt]
\textbf{Method} & \multicolumn{1}{c}{R-1} &
\multicolumn{1}{c}{R-2} &
\multicolumn{1}{c}{R-L} &
\multicolumn{1}{c}{Bertscore} &
\multicolumn{1}{c}{Avg-Drop}\\
\midrule
\textsc{Reviewer2} & 42.53 & 9.69 & 18.14 & 84.44 & 2.12\\
SEA-E & 47.45 & 12.09 & 19.04 & 84.93 & 2.12\\
DGE & \textbf{49.66} & \textbf{14.70} & \textbf{22.67} & \textbf{85.29} & \textbf{2.63}\\
SORT & 44.85 & 10.55 & 18.27 & 83.89 & 1.79\\
\midrule
\emph{TreeReview}\textsubscript{$-$\textsc{dec}} & 47.50 & 13.76 & 21.84 & 85.12 & 2.40\\
\emph{TreeReview}\textsubscript{$-$\textsc{exp}} & 47.65 & 13.75 & \textbf{22.03} & 85.12 & 2.39\\
\hdashline
\emph{TreeReview} & \textbf{49.94} & \textbf{14.24} & 21.78 & \textbf{85.27} & \textbf{2.57}\\

\bottomrule[1.2pt]
\end{tabular}
\caption{Results of ROUGE (R-1, R-2, and R-L) and BERTScore on the full review generation task. Avg-Drop denotes the average decrease of BERTScore when pairing generated reviews with reviews from other papers, reflecting the discriminability of BERTScore for this evaluation. Best scores are highlighted in bold.}
\label{tab:full_rouge_bert}
\end{table*}

\vspace{3pt}

\noindent\textbf{Results} \quad
The results are summarized in Table~\ref{tab:full_rouge_bert}. DGE and our \emph{TreeReview} achieve the highest scores for most metrics. Specifically, DGE attains the best R-1, R-2, R-L, and BERTScore among baseline methods, with \emph{TreeReview} closely matching or slightly exceeding these results, particularly on ROUGE-1 (49.94) and BERTScore (85.27). The \emph{TreeReview} ablations also demonstrate competitive performance, generally outperforming other baselines.

\vspace{3pt}

\noindent\textbf{Discussion} \quad
Despite these results, the differences between methods across ROUGE and BERTScore are marginal, with most values clustering within a narrow range. This is further highlighted by the Avg-Drop, which shows only a small decrease in BERTScore (ranging from 1.79 to 2.63) even when generated reviews are paired with unrelated reference reviews. These metrics primarily measure surface-level text overlap or semantic similarity, failing to capture the nuanced qualities of reviews. For instance, two reviews may differ significantly in their critical insights or actionable suggestions while still sharing similar phrasing or content overlap, leading to inflated scores that do not reflect true review quality. This limitation motivates our adoption of more sophisticated evaluation approaches, such as LLM-as-Judge and human evaluation, to capture the multifaceted nature of review quality.

\subsection{LLM-as-Judge Evaluation}
To evaluate the full review generation task, we employ an LLM-as-Judge approach, leveraging the \texttt{Gemini-2.5-Pro} to score system-generated reviews across multiple dimensions on a scale of 0-10. We design the following eight distinct quality dimensions:
\begin{itemize}[leftmargin=*]
    \item \textbf{Comprehensiveness}: Assesses whether the review covers all crucial aspects of the paper, such as the significance of the research problem, innovation, methodological soundness, etc.
    \item \textbf{Technical Depth}: Evaluates if the review demonstrates a strong understanding of the paper's technical content and the relevant research area.
    \item \textbf{Clarity}: Determines if the review clearly and accurately articulates the paper's strengths, weaknesses, and any points of confusion.
    \item \textbf{Constructiveness}: Assesses whether the review offers helpful and actionable suggestions that could genuinely aid in improving the paper.
    \item \textbf{Specificity}: Measures how focused the review is on particular issues within the given paper, rather than being generic or applicable to other papers.
    \item \textbf{Evidence Support}: Checks if the review substantiates its claims and feedback by referencing specific examples, sections, or data from the paper, and whether these references are faithful to the original content.
    \item \textbf{Consistency}: Evaluates the internal consistency of the review, ensuring it does not present contradictory statements or assessments.
    \item \textbf{Overall Quality}: Provides a holistic assessment of the review's quality, considering all the above dimensions.
\end{itemize}

The LLM judge is instructed to provide a concise textual justification for each score and output the assessment in a structured JSON format. The complete instructions, including the scoring scale descriptions, are provided in Fig.~\ref{fig:llm_as_judge_prompt}.

For each review, we conduct three independent scoring runs with the LLM at temperature 0.1 and average the scores across runs to obtain the final result. This multi-trial scheme helps to smooth out minor variance in LLM judges and improves reliability. We calculate the intraclass correlation coefficient (ICC) across independent scoring runs. The average-rater absolute ICC is 0.9642, indicating a high degree of consistency and robustness among LLM-based evaluations.

\subsection{Specificity Evaluation}
For evaluating the specificity of generated actionable feedback comments, we adopt the ITF-IDF metric proposed by \citet{du-etal-2024-llms}. This metric is reference-free and is designed to measure how specific and unique a review comment is to a particular paper, discouraging two undesirable scenarios: 1) repetitive segments within one review and 2) generic segments that appear across reviews for multiple papers. A higher ITF-IDF score indicates that the generated comments are more specific to the content of the individual paper and less generic across different papers. The ITF-IDF score is calculated as follows:

\begin{small}
\begin{equation}
\label{eq:itf_idf}
\text{ITF-IDF}=\frac{1}{W} \sum_{j=1}^{W} \left( \frac{1}{m_j} \sum_{i=1}^{m_j} \log \left(\frac{m_j}{O_{i}^{j}}\right) \times \log \left(\frac{W}{R_{i}^{j}}\right) \right)
\end{equation}
\end{small}

where $W$ represents the total number of papers in our dataset, $m_j$ is the number of generated feedback comments for paper $j$. $O_i^j$ measures the occurrence frequency of comment $i$ in paper $j$'s generated comments list (intra-paper occurrence), while $R_i^j$ measures the soft number of papers that also contain comment $i$ in their comments list (inter-paper occurrence). These components are calculated as:

\begin{small}
\begin{equation}
\label{eq:o_ij}
\setlength\abovedisplayskip{3pt}
\setlength\belowdisplayskip{3pt}
O_{i}^{j}=\sum_{k=1}^{m_{j}} \mathbb{I}\left(\text{sim}\left(c_{i}^{j}, c_{k}^{j}\right) \geq t\right) \cdot \text{sim}\left(c_{i}^{j}, c_{k}^{j}\right)
\end{equation}
\end{small}

\begin{small}
\begin{equation}
\label{eq:r_ij}
\setlength\abovedisplayskip{3pt}
\setlength\belowdisplayskip{3pt}
R_{i}^{j}=\sum_{l=1}^{w} \mathbb{I}\left(\max_{s} \text{sim}\left(c_{i}^{j}, c_{s}^{l}\right) \geq t\right) \cdot \max_{s} \text{sim}\left(c_{i}^{j}, c_{s}^{l}\right)
\end{equation}
\end{small}

where $c_i^j$ denotes the $i$-th comment in the generated comments list of paper $j$, $\text{sim}(\cdot,\cdot)$ denotes the semantic similarity between two comments. In this work, we implement it by encoding comments using \texttt{all-mpnet-base-v2}\footnote{\url{https://huggingface.co/sentence-transformers/all-mpnet-base-v2}} from SentenceBERT \cite{reimers-gurevych-2019-sentence} and calculating the cosine similarity. $t$ is a predefined similarity threshold (we set it to 0.5 in our experiments).

\subsection{Semantic Similarity-based Alignment}
To quantitatively evaluate the alignment between model-generated feedback comments and human reviewer comments, we adopt the semantic similarity-based metrics proposed by \citet{lou2024aaar}. We use SN-Precision, SN-Recall, and SN-F1 to measure the alignment between a single prediction list and multiple reference lists. 

Formally, given a generated comments list $p$ with $m$ comments, and reference comments lists $g^k$ from $r$ reviewers (where $g^k$ has $n_k$ comments for the $k$-th reviewer), the metrics are defined as follows:

\begin{small}
\begin{equation}
\begin{aligned}
    \text{SN-Precision} &= \frac{1}{m} \sum_{i=1}^{m} \left( \frac{1}{r} \sum_{k=1}^{r} \max_{j} \text{sim}(p_i, g_j^k) \right),
    \\ \text{SN-Recall} &= \frac{1}{r} \sum_{k=1}^{r} \left( \frac{1}{n_k} \sum_{j=1}^{n_k} \max_{i} \text{sim}(g_j^k, p_i) \right),
    \\ \text{SN-F1} &= 2 \cdot \frac{\text{SN-Precision} \cdot \text{SN-Recall}}{\text{SN-Precision} + \text{SN-Recall}}
\end{aligned}
\end{equation}
\end{small}

where $\text{sim}(\cdot, \cdot)$ denotes the semantic similarity between two comments. Again, we calculate the cosine similarity between the embeddings of comments encoded by \texttt{all-mpnet-base-v2}.

We use these metrics as a rough estimation of alignment in our experiments, as they rely on semantic distance in a latent space, which may not fully capture nuanced relationships between comment pairs. Additionally, the performance of the embedding model itself can also impact the accuracy of these metrics. As shown in Table \ref{tab:comments_result}, our \emph{TreeReview} framework achieves the highest SN-Precision (47.99\%) and a competitive SN-F1 (48.83\%), demonstrating strong semantic alignment with human feedback compared to baselines.

\subsection{LLM-based Alignment}
To provide a more nuanced and interpretable assessment of alignment beyond semantic similarity, we employ an LLM-based evaluation framework, following the methodology proposed by \citet{d2024marg}. This approach uses a powerful LLM to determine if a generated feedback comment conveys substantively the same meaning as a human-written reference comment. We utilize \texttt{Gemini-2.5-Pro} for this evaluation.

Given a set of generated feedback comments $C_{\text{gen}}$ and a set of reference human reviewer comments $C_{\text{real}}$ for a given paper, this approach aims to identify aligned comment pairs that convey the same critique or suggestion. The evaluation process involves two stages: (1) a many-to-many matching stage to identify candidate pairs across the full sets of comments, and (2) a pairwise evaluation stage to confirm alignments and assess their relatedness and relative specificity. In the first stage, we use \texttt{Gemini-2.5-Pro} to process both comment sets to propose potential matches. In the second stage, each candidate pair is individually evaluated to assign a relatedness score (``none'', ``weak'', ``medium'', or ``high'') and a relative specificity label (``less'', ``same'', or ``more'' for the generated comment compared to the reference). A pair is considered aligned if relatedness is rated as ``medium'' or ``high'' and the generated comment's specificity is ``same'' or ``more'' compared to the reference.

Using the aligned pairs, we compute the following metrics:
\begin{small}
\begin{equation}
\begin{aligned}
\text{Recall} &= \frac{|C_{gen} \overrightarrow{\cap} C_{real}|}{|C_{real}|}, \\
\text{Precision} &= \frac{|C_{gen} \overleftarrow{\cap} C_{real}|}{|C_{gen}|}, \\
\text{Pseudo-Jaccard} &= \frac{\text{intersection}}{|C_{gen}|+|C_{real}|-\text{intersection}}
\end{aligned}
\end{equation}
\end{small}
where $C_{gen}$ and $C_{real}$ represent the sets of generated and reference comments, respectively. The directional intersection operators $\overrightarrow{\cap}$ and $\overleftarrow{\cap}$ represent the set of aligned elements in the right or left operand, and the intersection is defined as $\frac{|C_{gen} \overrightarrow{\cap} C_{real}|+|C_{gen} \overleftarrow{\cap} C_{real}|}{2}$.

As mentioned in Appendix \ref{appendix:merge_comments}, we merged comments from all reviewers of each paper into a single reference set to establish a more comprehensive ground truth, which presents a more challenging evaluation scenario than comparing against individual reviewer comments. Additionally, we report the proportion of aligned comments rated as ``highly related'' and ``more specific'' to provide insight into the quality of matches. 

As reported in Table \ref{tab:comments_result}, our \emph{TreeReview} framework achieves the highest precision (32.10\%) and a competitive pseudo-jaccard score, demonstrating strong alignment with human reviewer feedback. \emph{TreeReview} also yields the highest proportion of ``highly related'' (15.72\%) and ``more specific'' comments among all methods, suggesting that it produces more precise and detailed feedback than baselines.

\subsection{Human Evaluation}
\label{appendix:human_eval_details}
To complement the automatic evaluation metrics and provide a more nuanced assessment of review quality, we conduct a comprehensive human evaluation. Here, we detail our evaluation protocol.

\vspace{3pt}

\noindent \textbf{Setup} \quad We recruit five expert evaluators (including three PhD candidates and two postdoctoral researchers) with significant experience in reviewing for top NLP and ML conferences or journals. They are recruited as volunteers through personal academic networks within the NLP research communities. To ensure a balanced evaluation, we sample a subset of 20 papers from our dataset, stratified by venue and acceptance decisions (statistics are shown in Table \ref{tab:dataset_statistics}). This subset was divided into 5 groups of 4 papers each for manageable workload distribution among evaluators.

\vspace{3pt}

\noindent \textbf{Procedure} \quad To ensure robust and unbiased assessments, we implement a two-round evaluation process. In the first round, each evaluator is assigned one unique group of 4 papers to assess. In the second round, the groups are shuffled and reassigned such that each evaluator reviews a different group, ensuring that every paper receives independent evaluations from two distinct evaluators. The evaluation was conducted in a pairwise comparison setup. For each paper, evaluators were presented with two anonymized outputs (either full reviews or lists of actionable feedback comments) generated by different methods. The order of presentation for the two outputs was randomized to mitigate order bias, and evaluators were blind to the identity of the methods that produced each output. 

Evaluators conduct the assessments via a web interface, as shown in Fig.~\ref{fig:human_evaluation_interface}. They are instructed to read the paper and carefully compare the pairs of reviews or feedback comments, selecting the superior output or indicating a tie based on the provided criteria. Evaluators are informed about the purpose of the study, how their evaluations would be used (i.e., for research purposes and potential publication), and that their identities would remain anonymous in all reports.

\vspace{3pt}

\noindent \textbf{Evaluation Criteria} \quad For the \emph{full review} task, evaluators are instructed to select the superior review based on criteria common in academic peer review: (1) \textit{thoroughness} (coverage of strengths, weaknesses, and key aspects such as originality, technical soundness, etc), (2) \textit{constructiveness} (actionable and helpful suggestions), (3) \textit{specificity} (degree to which comments are tailored to the paper rather than generic).
For the \emph{feedback comments} task, the focus was on: (1) \textit{accuracy} (correct identification of paper issues), (2) \textit{specificity}, and (3) \textit{helpfulness} (potential of comments to drive substantive improvement).

\vspace{3pt}

\noindent \textbf{(Pseudo) Win-Rate Calculation} \quad We calculate a ``pseudo win-rate'' for each method. Specifically, we aggregated the judgments from both evaluators for each paper-review pair. A method received a full win (+1) only when both evaluators independently judged it superior to the comparison method. When evaluators disagreed or both indicated a tie, each method received a half-point (+0.5). The final pseudo win-rate for each method against another method is calculated as the ratio of its accumulated points to the total number of comparisons. 

\vspace{3pt}

\noindent \textbf{Inter-evaluator Agreement} \quad
To assess evaluation reliability, we calculate both overall agreement (proportion of identical judgments) and the Cohen $\kappa$ coefficient. Our evaluation yielded an overall agreement of 0.75 and a Cohen $\kappa$ of 0.70, indicating substantial consistency among expert evaluators.

\section{Case Study}
\label{appendix:case_study}
To qualitatively assess the performance of \emph{TreeReview}, we conduct a case study on a sampled paper from our test set, focusing on the feedback comments generation task. 

Table \ref{tab:case_study_comments} presents a side-by-side comparison of feedback comments written by human reviewers (merged from multiple reviews) and the corresponding aligned comments generated by different methods. Comments are color-coded to indicate their degree of relatedness and specificity, as evaluated by the LLM-based alignment.

We observe that \emph{TreeReview} consistently produces feedback that is not only highly aligned with human comments but also demonstrates greater specificity and actionable insights compared to baselines. For example, while both MARG and SORT flag a lack of detail in the domain transformation process, \emph{TreeReview} further highlights missing parameter choices and missing justification for key design decisions, providing more concrete suggestions for revision. Similarly, on the issue of prior knowledge transfer, \emph{TreeReview} explicitly questions the suitability of mini-ImageNet as a source and calls for explanation of the relevance of features transferred\textemdash an aspect only vaguely touched upon by other methods. In scalability analysis, \emph{TreeReview} is the only method to directly critique the lack of computational complexity discussion in the paper, demonstrating its capacity for in-depth and targeted critique. In all instances, \emph{TreeReview}'s comments are both more specific and more closely aligned with the underlying concerns expressed by human reviewers, as indicated by the color-coded alignment assessments.

Fig.~\ref{fig:question_tree_example} illustrates a partial question tree constructed by \emph{TreeReview} for this paper, highlighting both top-down decomposed questions and dynamically expanded follow-up questions. The hierarchical decomposition enables comprehensive coverage across key review aspects (e.g., novelty, methodology, limitations), while the dynamic expansion mechanism allows the model to probe ambiguous or insufficiently addressed areas. We observe that many highly aligned comments trace back directly to these fine-grained and adaptively expanded questions. This hierarchical and adaptive questioning guides the LLM to systematically analyze the paper from multiple perspectives and granularities, contributing to the generation of specific and insightful feedback.

We also present the full review generated by \emph{TreeReview} for the sampled paper in Fig.~\ref{fig:full_review_example}, demonstrating its capability in producing comprehensive assessments.

\section{Further Analysis}
\subsection{Tree Structure Configuration Analysis}

To investigate the influence of the initial tree structure on the final review quality, we conduct an ablation study focusing on the same paper detailed in our case study (§\ref{appendix:case_study}). This analysis specifically evaluates the robustness of our framework to variations in the initial tree structure scale.

We compare the output of our standard setting against a configuration with a more constrained initial tree.
\begin{itemize}
    \item \textbf{Original Tree:} The initial tree generation allows for a maximum of 5 sub-questions at the first level ($W^{1}_{\max} = 5$), with subsequent levels having maximums of 4 and 3 sub-questions respectively.
    \item \textbf{Compact Tree:} We reduce the maximum number of sub-questions for the first level to 4 ($W^{1}_{\max} = 4$), with subsequent levels having maximums of 3 and 2.
\end{itemize}

For both configurations, the maximum number of follow-up questions generated by the dynamic expansion mechanism remains constant ($W^{\text{exp}}_{\max} = 2$).

Our analysis reveals that the dynamic expansion mechanism provides significant robustness, effectively compensating for potential omissions in a less extensive initial tree. For instance, in the case study paper:
\begin{itemize}
    \item With the \textbf{Original Tree}, the initial question decomposition included an inquiry about dataset statistics: \hl{\textit{``What is the distribution of features (numerical, categorical, missing values) in the datasets used...?''}}
    \item With the \textbf{Compact Tree}, this specific question was absent from the initial static structure. However, the dynamic expansion mechanism subsequently generated a semantically equivalent inquiry during the review process: \hl{\textit{``What are the key characteristics (e.g., number of columns, data types, distributions, and presence of outliers) of each dataset used...?''}}
\end{itemize}

This result demonstrates that \emph{TreeReview} is \textbf{robust to variations in the initial tree configuration}. The dynamic expansion mechanism ensures that critical aspects of the paper are thoroughly examined, even if they are not captured during the initial question decomposition phase. This enhances the reliability of our framework by reducing its sensitivity to the hyperparameters governing the initial tree structure.

\subsection{Error Mode Analysis}

To better understand the limitations of our proposed method, we conduct a qualitative analysis of samples that scored low in the LLM-based alignment evaluation. This process reveal several recurring patterns in failure cases:
\begin{itemize}
    \item \textbf{Multimodal Content:} Our current implementation does not process multimodal inputs. Consequently, it cannot capture feedback related to visual elements. For example, human reviewer comments like \textit{``Xsync notation inconsistency, especially in Figure 1''} fall outside the scope of our method’s capabilities. 
    
    \item \textbf{Fine-Grained Consistency:} \emph{TreeReview} can be challenged by subtle inconsistencies that span different sections of a paper. For instance, in one case, the ground truth feedback pointed out confusion between \textit{``Equation (4) and the distinction between \(p(Q|X^*,Z)\) and \(p(Q|Z)\).''} Correctly identifying such issues requires tracking detailed information scattered across the manuscript. Although our final aggregation step is designed to synthesize a holistic review, ensuring complete consistency at this granular level remains difficult.
\end{itemize}

\subsection{Performance Analysis by Paper Type}
\label{sec:appendix_performance_by_type}

To investigate whether the model’s performance correlates with the type of document being reviewed, we categorize the papers in our benchmark into five types: Method/Algorithm, Application, Theoretical, Analysis, and Benchmark. We then analyze the relationship between document type and performance, using the Jaccard similarity of actionable feedback comments as the primary performance indicator.

Our statistical tests across the full dataset showed \textbf{no significant correlation} between paper type and alignment performance. However, to probe for more subtle trends, we isolate the lowest-performing 20\% of papers based on their alignment scores. In this subset, we observe that \textbf{Application} and \textbf{Theoretical} papers are slightly overrepresented compared to their overall proportions in the full dataset.

We hypothesize that this may be due to the unique demands of these paper types:
\begin{itemize}
    \item \textbf{Application papers} often focus on specific practical scenarios that require diverse, domain-specific context, which can be more difficult for an LLM to align with perfectly.
    
    \item \textbf{Theoretical papers} frequently contain intricate logical reasoning and abstract concepts, challenging the LLM's ability to generate precisely aligned technical feedback.
\end{itemize}

\section{Prompts used in \emph{TreeReview}}
\label{appendix:treereview_prompts}

We present all prompts utilized in the \emph{TreeReview} framework in the following figures: Fig.~\ref{fig:prompt_question_generator}, Fig.~\ref{fig:prompt_answer_leaf}, Fig.~\ref{fig:prompt_answer_intermediate}, Fig.~\ref{fig:prompt_answer_root_full}, and Fig.~\ref{fig:prompt_answer_root_comments}. These prompts guide the various stages as described in the methodology.

\begin{figure*}[t]
    \centering
    \small
    \begin{tcolorbox}[colframe=black, colback=white, width=\textwidth, boxrule=0.2mm]
    You are a highly experienced area chair for top-tier academic conferences. Your task is to assess the quality of a review for a given paper based on specific evaluation criteria. You must ensure your assessment is professional, objective, and well-reasoned.
    \\ \hspace*{\fill} \\
    You will receive an academic paper and its associated peer review. 
    \\ \hspace*{\fill} \\
Firstly, take time to thoroughly read and understand both the paper and its review.
\\ \hspace*{\fill} \\
Then, analyze and score the quality of the review based on specific criteria outlined below:

1. **Comprehensiveness**: Does the review assess all important dimensions of the paper, including the significance of the research question, innovation and originality, methodological rigor, experimental design and analysis, potential impact on the field, and other key aspects?

2. **Technical Depth**: Does the review demonstrate a thorough understanding of the paper's content and the related research domain? Does it identify subtle yet significant technical issues?

3. **Clarity**: Does the review accurately and clearly identify specific strengths, weaknesses, and unclear aspects of the paper? 

4. **Constructiveness**: Is the review constructive and helpful in nature? Can the provided suggestions or insights really help improve the paper?

5. **Specificity**: Is the review focused on particular issues within the given paper, rather than being overly generic or applicable to other papers?

6. **Evidence Support**: Does the review reference specific examples, sections, or data from the paper to substantiate its observations and feedback? Is the referenced content faithful to the original paper?

7. **Consistency**: Is the review internally consistent? Does it contain contradictory viewpoints?

8. **Overall Quality**: Considering all aspects, how would you score the overall quality of the review?
\\ \hspace*{\fill} \\
Before assigning any scores, carefully analyze the review against each evaluation criterion, thinking step-by-step.
\\ \hspace*{\fill} \\
For each criterion, first provide a concise reason, then assign a score using the following scale:

- 0-2: Severely deficient - Fails to meet basic standards

- 3-4: Below acceptable standards - Major improvements needed

- 5-6: Acceptable - Meets minimum standards but has clear limitations

- 7-8: Good - Exceeds standard expectations with minor limitations

- 9-10: Excellent - Exemplary quality with minimal or no limitations
\\ \hspace*{\fill} \\
Format your assessment as a JSON object with the following structure:
\begin{lstlisting}[basicstyle = \ttfamily]
{
  "Comprehensiveness": {
    "reason": str,
    "score": int
  },
  
  ...
  
  "Overall Quality": {
    "reason": str,
    "score": int
  }
}
\end{lstlisting}

\hspace*{\fill} \\
Only output the final JSON object.
    \end{tcolorbox}
    \caption{Instructions for the LLM-as-Judge evaluation.}
    \label{fig:llm_as_judge_prompt}
\end{figure*}

\begin{figure*}[htbp!]
    \centering
    \includegraphics[width=\textwidth, height=0.9\textheight, keepaspectratio]{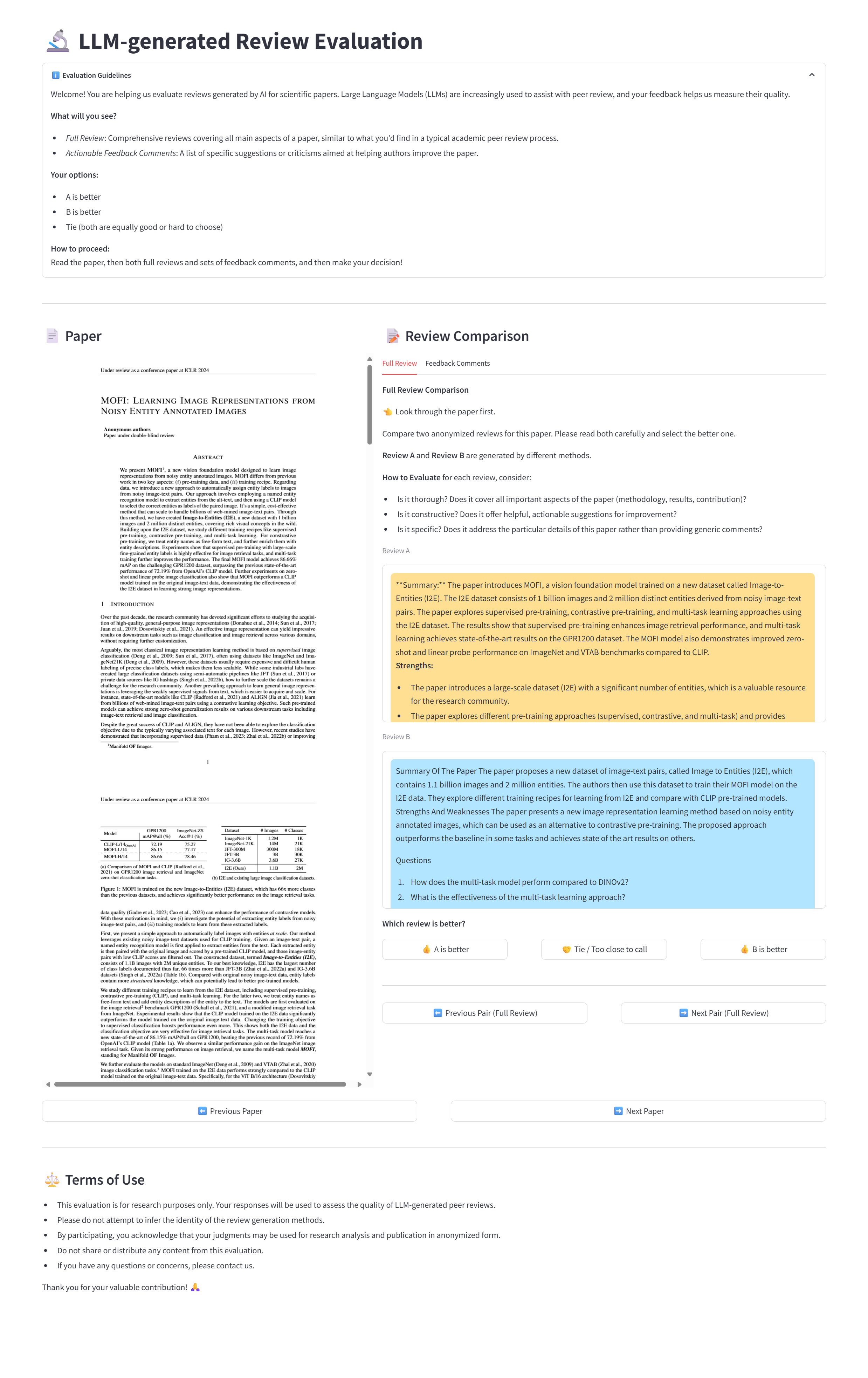}
    \caption{The interface used for human evaluation. It includes guidelines for evaluators, the paper PDF, and pairs of full reviews or sets of feedback comments for pairwise comparison.}
    \label{fig:human_evaluation_interface}
\end{figure*}

\definecolor{relhigh_specmore}{RGB}{198,239,206}
\definecolor{relhigh_specsame}{RGB}{189,215,238}
\definecolor{relmed_specmore}{RGB}{189,215,238}
\definecolor{relmed_specsame}{RGB}{255,199,206}

\begin{table*}[ht]
\centering
\renewcommand{\arraystretch}{1.5}
\setlength{\tabcolsep}{4pt}
\scriptsize 

\begin{tabular}{>{\raggedright\arraybackslash}m{0.36\textwidth} !{\vrule width 1pt} >{\centering\arraybackslash}m{0.08\textwidth} !{\vrule width 1pt} >{\raggedright\arraybackslash}m{0.50\textwidth}}
\toprule[1.2pt]
\textbf{Human Reviewer Comment} & \textbf{Method} & \textbf{Aligned Generated Comment} \\
\midrule
\multirow{3}{=}{
The description of the domain transformation, a key part of the paper, lacks sufficient underlying intuition, motivation, justification, technical correctness, and clarity. Provide more details (e.g., figures, equations, or pseudo-code) for all model parts, especially the domain transformation, which was hard to understand. Clarify if alternatives to tiling (like resizing or padding) were considered for matrix handling and why tiling was chosen.
} 
    & SORT & 
    \cellcolor{relhigh_specsame}The description of the domain transformation process is not sufficiently detailed. $\cdots$ The motivation for using a three-channel image representation is not well-explained. $\cdots$\\
    \cline{2-3}
    & MARG &
    \cellcolor{relhigh_specsame}The description of the proposed TabLEye method lacks key details regarding the transformation of tabular data into the image domain $\cdots$ The authors should provide a more detailed explanation of the data transformation process and the mechanisms for incorporating prior knowledge $\cdots$\\
    \cline{2-3}
    & Ours &
    \cellcolor{relhigh_specmore}The domain transformation process, detailed in Section 3.2 and Algorithm 1 (Appendix D), lacks crucial implementation details $\cdots$ The specific values or ranges for parameters like \(n_r\), \(n_c\), and the elements within the matrices $R$ and $Q$ are not provided. The justification for choosing a final image size of (84, 84) is also missing $\cdots$\\
\midrule

\multirow{2}{=}{
Further justification and detail are needed regarding the domain transformation: Why would the proposed ranking and rearrangement of feature/pixel distances resemble information in natural images? Provide intuition for why a network pretrained on natural images (like miniImageNet) should be useful for these artificially created tabular images and how the potential domain gap is overcome (beyond the t-SNE visualization).}
    & DPW & 
    \cellcolor{relmed_specsame}2. The domain transformation method seems arbitrary and lacks a clear explanation of why the specific distance metrics and spatial arrangement are chosen. The impact of the $\alpha$ parameter in the distance calculation is not discussed.\\
    \cline{2-3}
    & Ours &
    \cellcolor{relhigh_specmore}$\cdots$ The relevance of mini-ImageNet, a dataset of natural images, to tabular data is not clearly established. While the t-SNE visualization in Fig.~4 suggests some overlap between the feature spaces of natural images and tabular images, a more detailed explanation is needed to justify the choice of mini-ImageNet as a source of prior knowledge. The authors should explain what specific features or patterns learned from mini-ImageNet are relevant to tabular data and why this dataset is more suitable than other potential sources of prior knowledge.\\

\midrule

\multirow{2}{=}{
The description of experimental details lacks clarity, making it difficult to judge the results. Specifically for Table 1 (demonstrating the benefit of prior knowledge): Clarify the exact experimental setting used for the comparisons. Are the experiments 'without image-pretraining' simply trained on the generated tabular images using a randomly initialized backbone? Are the image-pretrained methods further fine-tuned on some tabular image data? This information is needed to assess knowledge transfer and overfitting risks. 
} 
    & DPW & 
    \cellcolor{relmed_specsame} 7. The experimental results are only briefly mentioned ("superior performance by outstripping the TablLMO in a 4-shot task with a maximum 0.11 AUC and a STUNT in a 1-shot setting, where it led on average by $3.17\%$ accuracy"). There is no detailed experimental setup, datasets used, evaluation metrics, or statistical significance analysis provided in this excerpt.\\
    \cline{2-3}
    & Ours & 
    \cellcolor{relmed_specmore} The 'No Img' condition uses a randomly initialized backbone trained on tabular images. It's unclear if this backbone is trained to convergence or for the same number of epochs as the 'Img' condition. If the 'No Img' backbone is not adequately trained, the comparison might be biased, as the performance difference could be attributed to insufficient training rather than the absence of prior knowledge. Clarifying the training protocol for the 'No Img' backbone is crucial for the validity of the ablation study.\\

\midrule

\multirow{2}{=}{
The method may face scalability issues with high-dimensional tabular data, as converting tables with many features into images can result in impractically large image dimensions, hindering scalability and efficiency. Consideration should be given to how the architecture (CNN or alternatives) could be adapted for such datasets while maintaining computational efficiency.
} 
    & MARG & 
    \cellcolor{relmed_specmore} The paper uses a fixed image size of 84x84 to represent tabular data as images. However, the paper does not adequately address the potential limitations of this approach when dealing with datasets with varying numbers of features. Specifically, it is unclear how the method handles datasets with a large number of features. If the 84x84 image size is insufficient to represent all features without significant information loss, this could negatively impact performance. The authors should investigate and discuss the impact of feature number on performance, potentially by experimenting with different image sizes or feature selection techniques on datasets with a high number of features.\\
    \cline{2-3}
    & Ours & 
    \cellcolor{relhigh_specmore}The discussion section fails to adequately address the computational complexity and scalability of TabLEye. While the paper mentions that TabLEye has a smaller model size than TabLLM (Section 4.3), it doesn't provide a comprehensive analysis of the computational cost associated with the domain transformation process, especially for high-dimensional tabular datasets. Furthermore, the scalability of TabLEye to larger datasets is not discussed. Addressing these aspects is essential for understanding the practical applicability of TabLEye in real-world scenarios.\\

\midrule

\multirow{2}{=}{
Establishing meaningful spatial relationships within the transformed images can be challenging for heterogeneous tabular data, potentially limiting the method's applicability and suggesting it may not be a universally applicable solution for all tabular learning problems, especially those with highly diverse data structures.
} 
    & MARG & 
    \cellcolor{relmed_specmore} The paper lacks an explicit discussion of how TabLEye handles the challenges posed by statistical data heterogeneity across different tabular datasets. While the experimental results showcase performance on various datasets, and Section 4 mentions the diversity of features, there is no analysis of how the statistical properties of these datasets differ and how TabLEye is designed to be robust to such variations. Addressing this point is important because real-world tabular data often exhibits significant statistical heterogeneity, and understanding TabLEye's ability to handle this is crucial for assessing its practical applicability. \\
    \cline{2-3}
    & Ours & 
    \cellcolor{relhigh_specmore} The paper's core assumption that feature similarity, as measured by Euclidean distance, can be effectively translated into spatial relationships in an image is not sufficiently justified. The paper states, "We hypothesize that the difference between images and tabular data lies in the association with neighboring values and spatial relations" (Section 3.2), but it doesn't provide a strong rationale for why this specific type of spatial encoding is universally suitable for tabular data, especially considering the heterogeneity of tabular datasets. A more detailed explanation, possibly with illustrative examples or a theoretical analysis, is needed to support this central hypothesis.\\

\bottomrule[1.2pt]
\end{tabular}
\caption{
\textbf{Case study of aligned comments generated by different methods for a sample paper.} The table shows human reviewer comments and their corresponding generated comments from various methods. Color coding indicates the alignment assessment: \colorbox{relmed_specsame}{medium relatedness, same specificity}, \colorbox{relmed_specmore}{medium relatedness, more specificity}, \colorbox{relhigh_specsame}{high relatedness, same specificity}, and \colorbox{relhigh_specmore}{high relatedness, more specificity}.
}
\label{tab:case_study_comments}
\end{table*}

\begin{figure*}[htbp!]
    \centering
    \includegraphics[width=\textwidth, height=0.9\textheight, keepaspectratio]{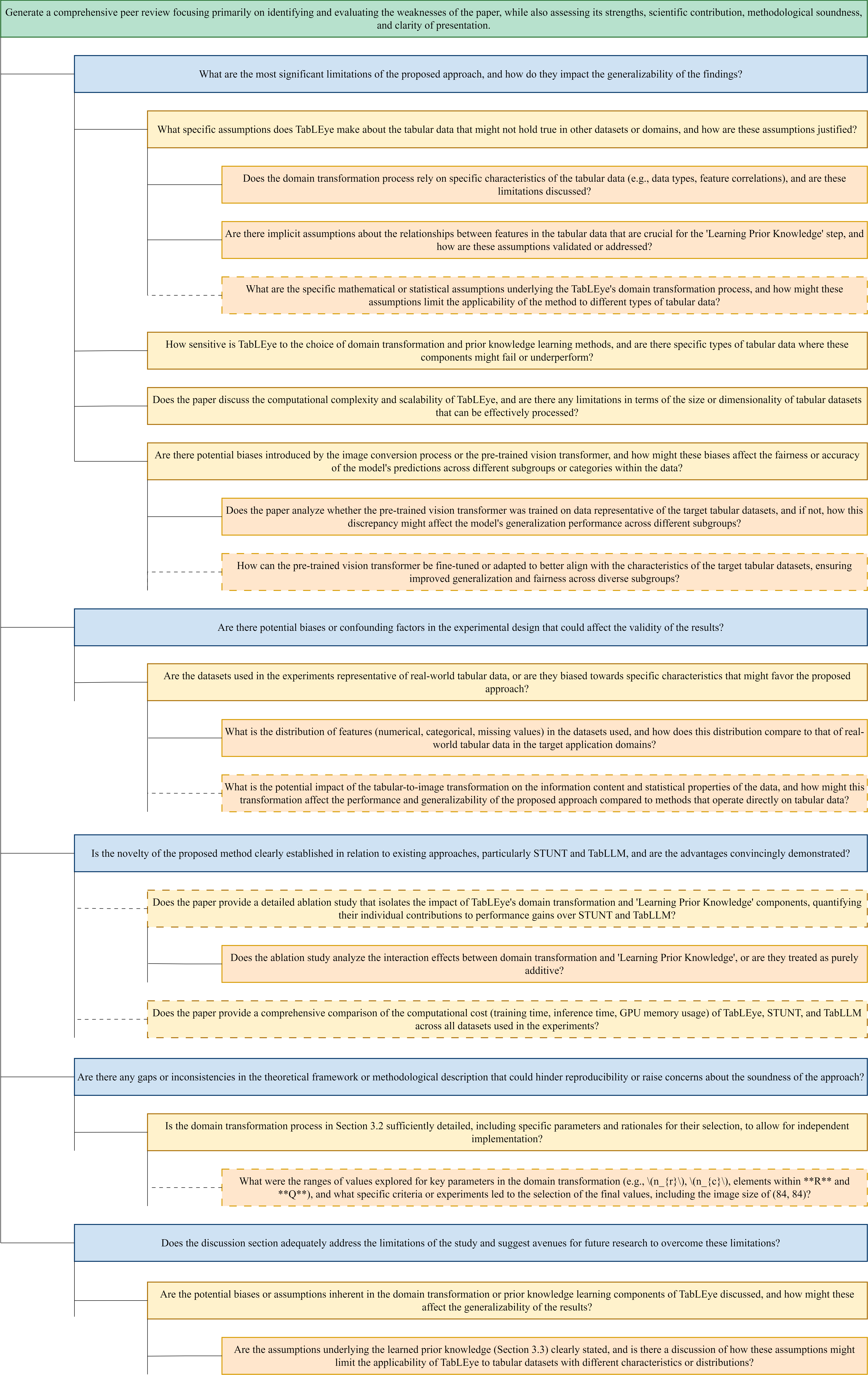}
    \caption{Example of review question tree constructed by \emph{TreeReview}. Only a subset of key questions is shown due to the large tree size. Different colors indicate distinct question depths, while dynamic expansion processes and newly generated follow-up questions are highlighted by dashed lines.}
    \label{fig:question_tree_example}
\end{figure*}

\begin{figure*}[t]
    \centering
    \scriptsize
    \begin{tcolorbox}[colframe=black, colback=white, width=\textwidth, boxrule=0.2mm]
    \textbf{Summary:} 
    \\ \hspace*{\fill} \\
    This paper introduces TabLEye, a novel framework for few-shot tabular learning. TabLEye transforms tabular data into images, leveraging pre-trained image models for prior knowledge and few-shot classification. The paper claims TabLEye overcomes limitations of existing methods like TabLLM and STUNT, achieving superior performance with less data and computation. Experiments on public datasets are presented to validate the approach. 
    \\ \hspace*{\fill} \\
    \textbf{Strengths:}
    \begin{itemize}
    \setlength{\itemsep}{0pt}
    \setlength{\parsep}{0pt}
    \setlength{\parskip}{0pt}
    \item The idea of using image-based prior knowledge for tabular learning is novel and potentially valuable. 
    \item The paper identifies limitations of existing tabular learning methods, such as the need for large datasets or specific feature types. 
    \item The experimental results show promising performance compared to baselines in some scenarios. 
    \item The paper is well-structured and clearly written.
    \end{itemize}
    \textbf{Weaknesses:}
    \begin{itemize}
    \setlength{\itemsep}{0pt}
    \setlength{\parsep}{0pt}
    \setlength{\parskip}{0pt}
    \item \textbf{Unjustified Assumptions in Domain Transformation:} The core of TabLEye relies on the assumption that feature similarity can be effectively represented by Euclidean distance and that aligning feature and pixel distances preserves essential relationships. This is a strong assumption that lacks explicit justification. Datasets with non-linear feature relationships or where Euclidean distance is not a suitable measure of similarity would likely suffer. The paper needs to provide a more robust justification for this choice or explore alternative similarity measures.
    \item \textbf{Lack of Sensitivity Analysis:} The paper lacks an analysis of TabLEye's sensitivity to the choice of domain transformation and prior knowledge learning methods. Without ablation studies or experiments demonstrating performance degradation on specific datasets, it's impossible to determine the robustness of the approach. For example, how does the performance change with different image sizes or different pre-trained image models?
    \item \textbf{Missing Computational Complexity Analysis:} The paper does not provide a direct analysis of TabLEye's computational complexity or scalability, despite implicitly addressing feature dimensionality through a fixed-size image transformation. The lack of empirical evidence or theoretical analysis to support claims of scalability represents a significant gap. A comparison of training and inference times with STUNT and TabLLM is crucial.
    \item \textbf{Potential Biases Not Addressed:} The paper fails to address potential biases introduced by the image conversion process or the pre-trained vision transformer. The absence of discussion on data encoding biases, analysis of the pre-trained vision transformer's training data, and experiments to evaluate fairness across different subgroups raises concerns about the model's potential for biased predictions and its applicability across diverse populations.
    \item \textbf{Questionable Representativeness of Datasets:} The paper lacks detailed descriptions of feature distributions and how they compare to real-world tabular data, raising concerns about generalizability. The tabular-to-image transformation also introduces potential information loss and alterations to statistical properties, despite the ablation study.
    Inadequate Control for Confounding Variables: The experimental setup inadequately controls for confounding variables. The paper fails to address or quantify class imbalance and lacks details on dataset sizes in the ablation study.
    \item \textbf{Missing Hyperparameter Tuning Details:} The lack of detailed information regarding the hyperparameter tuning process for the baseline methods (TabLLM, STUNT, XGB, and TabNet) raises concerns about biased comparisons. The absence of this information casts doubt on whether the baselines were optimally configured.
    \item \textbf{Insufficient Detail for Reproducibility:} The domain transformation process lacks sufficient detail for independent implementation. The paper fails to provide specific values or ranges for key parameters (e.g., ($n_{r}$), ($n_{c}$), elements within $R$ and $Q$), nor does it justify the final image size of (84, 84). The computational complexity of the domain transformation is also not discussed. Furthermore, Section 3.3's description of prior knowledge learning is incomplete, lacking explicit mathematical equations and a clear articulation of the underlying assumptions and their limitations, especially in the context of tabular data.
    \item \textbf{Discussion Section Lacks Depth:} The discussion section of the paper inadequately addresses the limitations of the proposed TabLEye approach and fails to provide concrete directions for future research.
    \end{itemize}
    \textbf{Questions:}
    \begin{itemize}
    \setlength{\itemsep}{0pt}
    \setlength{\parsep}{0pt}
    \setlength{\parskip}{0pt}
    \item Can you provide a more detailed justification for using Euclidean distance as a measure of feature similarity in the domain transformation process? Are there alternative similarity measures that might be more appropriate for certain types of tabular data?
    \item Can you provide a sensitivity analysis of TabLEye's performance with respect to different image sizes, pre-trained image models, and other key hyperparameters?
    \item Can you provide a detailed analysis of the computational complexity and scalability of TabLEye, including a comparison of training and inference times with STUNT and TabLLM?
    \item How does TabLEye address potential biases introduced by the image conversion process or the pre-trained vision transformer? Can you provide an analysis of the pre-trained vision transformer's training data and experiments to evaluate fairness across different subgroups?
    \item Can you provide more detailed descriptions of the feature distributions of the datasets used in the experiments and how they compare to real-world tabular data?
    \item Can you provide more details on the hyperparameter tuning process for the baseline methods?
    \item Can you provide more specific details on the domain transformation process, including the values or ranges for key parameters and a justification for the final image size of (84, 84)?
    \item Can you provide a more complete description of the prior knowledge learning process, including explicit mathematical equations and a clear articulation of the underlying assumptions and their limitations?
    \item Can you provide a dedicated discussion section outlining specific, actionable future research directions that build upon the TabLEye framework to address its identified limitations?
    \end{itemize}
    \textbf{Soundness:} 3 
    \\ \hspace*{\fill} \\
    \textbf{Presentation:} 3 
    \\ \hspace*{\fill} \\
    \textbf{Contribution:} 2 
    \\ \hspace*{\fill} \\
    \textbf{Rating:} 5 
    \\ \hspace*{\fill} \\
    \textbf{Confidence:} 4
    \end{tcolorbox}
    \caption{Example of full review produced by \emph{TreeReview}.}
    \label{fig:full_review_example}
\end{figure*}

\begin{figure*}[t]
    \centering
    \small
    \begin{tcolorbox}[colframe=black, colback=white, width=\textwidth, boxrule=0.2mm]
       You are an expert in academic peer review, specializing in decomposing high-level review questions into structured, critical sub-questions that help reviewers thoroughly evaluate a paper. You will receive the metadata of the submitted paper (title, abstract, table of contents) and a parent review question. Your task is to generate sub-questions that are specific, actionable, and focused on distinct aspects of the parent question, following MECE principles (Mutually Exclusive, Collectively Exhaustive).
        \\ \hspace*{\fill} \\
        TASK REQUIREMENTS:
        
        1 Contextual Awareness:
        
        - You are a reviewer tasked with evaluating the paper. Your questions should reflect a critical and analytical perspective, aimed at identifying strengths, weaknesses, and areas that require further clarification or improvement.
        
        - At the root level (Current Depth in Review Tree: 0), generate sub-questions that cover the major aspects of a peer review, such as novelty, quality, clarity, significance, etc.
        
        - At deeper levels, generate increasingly specific sub-questions that probe finer details of the paper’s content.
        
        - If the parent question is already sufficiently detailed and does not require further decomposition, return an empty list.
        
        2 Question Quality:
        
        - Ensure sub-questions are:
        
        -- Mutually Independent: No overlap between sub-questions.
        
        -- Collectively Exhaustive: Together, they cover all key aspects of the parent question.
        -- Locally Answerable: Try to ensure that sub-questions can be answered by reading fragments of the paper (specific sections, paragraphs, or technical elements), so that the reviewer can focus their attention on specific content of the paper.
        
        -- Paper Specific: Contextualize sub-questions within the paper’s research content.
        
        - Generate the minimum number of sub-questions necessary to thoroughly address the parent question, while ensuring that each question is critical, specific, and contributes meaningfully to the evaluation. Avoid generating redundant or overly granular questions unless absolutely necessary.
        
        - Maintain scientific rigor and focus on critical evaluation, avoiding superficial or overly broad questions.
        
        3 Peer-Review Focus:
        
        - Frame questions from the perspective of a reviewer, not the author. For example:
        Instead of asking, "Does the author explain the methodology clearly?" ask, "Is the methodology described in sufficient detail to allow for reproducibility?"
        
        4 Question Scope:
        
        - Focus solely on textual components of the paper, excluding figures, tables, or visual elements from consideration.
        
        5 Number of sub-questions:
        
        - Generate up to \{QUESTIONS NUM\} sub-questions.
        
        - If the parent question is already sufficiently detailed, return empty array.
        \\ \hspace*{\fill} \\
        INPUT:
        
        - Paper Title: \{PAPER TITLE\}
        
        - Paper Abstract: \{PAPER ABSTRACT\}
        
        - Paper Table of Contents: \{PAPER TOC\}
        
        - Current Depth in Review Tree: \{NODE DEPTH\}
        
        - Parent Question: \{PARENT QUESTION\}
        \\ \hspace*{\fill} \\
        OUTPUT FORMAT:
        
        A JSON array of strings containing up to \{QUESTIONS NUM\} sub-questions.
        
        Example: ["Question1", "Question2", "Question3"]
        
        If no further sub-questions are needed, return an empty JSON array: []
        
        Only output the JSON array. 
    \end{tcolorbox}
    \caption{Prompt for the Question Generator in decomposing review questions.}
    \label{fig:prompt_question_generator}
\end{figure*}

\begin{figure*}[t]
    \centering
    \small
    \begin{tcolorbox}[colframe=black, colback=white, width=\textwidth, boxrule=0.2mm]
        You specialize in providing precise, evidence-based answers to review questions for submitted paper. You operate at the leaf-node level of a peer-review question tree. Your answers will directly support higher-level critique synthesis.
        \\ \hspace*{\fill} \\
    TASK REQUIREMENTS:
    
    1. Only use information explicitly stated in the provided Relevant Context.
    
    2. Avoid making inferences, predictions, or hypotheses that are not directly supported by the text. If the text is ambiguous or incomplete, acknowledge the limitation and refrain from filling gaps with assumptions.
    
    3. Use formal, precise, and objective language. Avoid casual phrasing, exaggeration, or emotional language.
    
    4. Provide Detailed Evidence: For each comment, include specific evidence from the given context (e.g., quotes, section references, or data points) to justify your point.
    \\ \hspace*{\fill} \\
    INPUT:
    
    - Review Question: \{QUESTION\}
    
    - Relevant Context: \{CONTEXT\}
    \\ \hspace*{\fill} \\
    OUTPUT FORMAT: 
    
    A single string containing only the answer to the review question.
    \\ \hspace*{\fill} \\
    Your final answer:
    \end{tcolorbox}
    \caption{Prompt for the Answer Synthesizer in answering leaf questions.}
    \label{fig:prompt_answer_leaf}
\end{figure*}

\begin{figure*}[t]
    \centering
    \small
    \begin{tcolorbox}[colframe=black, colback=white, width=\textwidth, boxrule=0.2mm]
    As an intermediate node in the peer review question tree, your role is to analyze and synthesize answers from sub-questions (child nodes) to determine whether the evidence is sufficient to address the current node's question. Your primary goal is to evaluate the paper from a critical reviewer's perspective, identifying strengths, weaknesses, and potential gaps in the research. Based on the provided sub-questions and answers, you must first determine whether the evidence is sufficient to address the main question. If sufficient, synthesize a critical review segment for your parent node; if insufficient, propose additional questions to deepen the investigation. Your output must bridge lower-level evidence to higher-level evaluations, ensuring the review process is both rigorous and logically structured.
    \\ \hspace*{\fill} \\
    INSTRUCTION:
    
    If the evidence is sufficient to address the main question, follow the "Sufficient Evidence" task requirements and output format.
    
    If the evidence is insufficient to address the main question, follow the "Insufficient Evidence" task requirements and output format.
    \\ \hspace*{\fill} \\
    TASK REQUIREMENTS FOR SUFFICIENT EVIDENCE:
    
    1. Critical Reviewer Perspective: From the perspective of a peer reviewer, not the author. Focus on evaluating the paper's claims, methodology, and conclusions critically. Avoid defending the paper or emphasizing its contributions without sufficient evidence.
    
    2. Input-Bound Synthesis: Use only the provided sub-Q\&A pairs. Never reference external knowledge or invent claims.
    
    3. Analytical Depth: Dive deeply into the sub-answers to uncover patterns, contradictions, and gaps. Synthesize insights that go beyond surface-level observations, critically evaluating the strength of evidence and exploring the broader implications of the findings.
    
    4. Critical Thinking: Consider the implications of the sub-answers and how they collectively address the main question. Highlight any significant findings or unresolved issues.
    
    5. Provide Detailed Evidence: For each insight in your synthesized answer, include specific evidence from the sub-Q\&A pairs (e.g., quotes, section references, or data points) to justify your point.
    
    6. Chain of Thought: Clearly articulate your reasoning process, showing how you derived your conclusions from the sub-answers. This should include a step-by-step explanation of your thought process.
    \\ \hspace*{\fill} \\
    OUTPUT FORMAT FOR SUFFICIENT EVIDENCE:
    
    A JSON object containing the chain of thought and the synthesized answer.
    
    Use the following JSON schema and ensure proper escaping of special characters (e.g., double quotes, forward/backward slashes, etc):
    
   \{
   
        "chain\_of\_thought": str,
        
        "synthesized\_answer": str
        
    \}
    \\ \hspace*{\fill} \\
    TASK REQUIREMENTS FOR INSUFFICIENT EVIDENCE:
    
    1. Evidence Assessment: If the provided sub-Q\&A pairs are insufficient to answer the main question, propose up to {MAX QUESTION NUM} follow-up questions that need to be answered to address the main question adequately.
    
    2. Analytical Depth: Analyze the sub-answers to identify specific areas where the evidence is lacking or contradictory. Determine what additional information is required to address the main question adequately.
    
    3. Chain of Thought: Clearly articulate your reasoning process, showing how you identified the gaps in the evidence and why the proposed follow-up questions are necessary. This should include a step-by-step explanation of your thought process.
    \\ \hspace*{\fill} \\
    OUTPUT FORMAT FOR INSUFFICIENT EVIDENCE:
    
    A JSON object containing the chain of thought and up to {MAX QUESTION NUM} follow-up questions.
    
    Use the following JSON schema and ensure proper escaping of special characters (e.g., double quotes, forward/backward slashes, etc):
    
    \{
    
      "chain\_of\_thought": str,
      
      "follow\_up\_questions": list[str]
      
    \}
    \\ \hspace*{\fill} \\
    INPUT:
    
    - Question: {QUESTION}
    
    - Sub-questions and answers: {QUESTIONS AND ANSWERS}
    \\ \hspace*{\fill} \\
    Only output the JSON object.
    \end{tcolorbox}
    \caption{Prompt for the Answer Synthesizer in aggregating answers and generating follow-up questions for intermediate questions.}
    \label{fig:prompt_answer_intermediate}
\end{figure*}

\begin{figure*}[t]
    \centering
    \small
    \begin{tcolorbox}[colframe=black, colback=white, width=\textwidth, boxrule=0.2mm]
    You are an expert reviewer tasked with providing a thorough, critical, and constructive review for a scientific paper submitted for publication. A review aims to determine whether a submission will bring sufficient value to the community and contribute new knowledge. You will be given the full paper content and a set of question-answer pairs about the paper, which are obtained through in-depth understanding and analysis of the paper. These Q\&A pairs will be very helpful for you to build a high-quality review. Please follow the instructions and requirements provided below:
    \\ \hspace*{\fill} \\
    INSTRUCTIONS
    
    1. Firstly, you should carefully read through the entire paper.
    
    2. Secondly, it’s important to use the questions and their corresponding answers as a guiding framework to help you deeply understand the paper and ensure a comprehensive review.
    
    3. Based on the analysis from the first two steps, compose a thorough and comprehensive review.
    \\ \hspace*{\fill} \\
    REQUIREMENTS
    
    1. While the question-answer pairs are important inputs for your analysis, your review should focus on the paper itself and avoid directly mentioning the Q\&A pairs. Instead, use the insights from them to inform your review process.
    
    2. In your review, you must cover the following aspects:
    
    [ICLR and NIPS Reviewer Guideline]
    \\ \hspace*{\fill} \\
    INPUT
    
    - Paper Content: {PAPER CONTENT}
    
    - Questions and answers: {QUESTIONS AND ANSWERS}
    \\ \hspace*{\fill} \\
    OUTPUT FORMAT
    
    Here is the template for a review format. You must follow this format to output the integrated review results:
    
    **Summary:**
    
    Summary content
    
    **Strengths:**
    
    Strengths result
    
    **Weaknesses:**
    
    Weaknesses result
    
    **Questions:**
    
    Questions result
    
    **Soundness:**
    
    Soundness result
    
    **Presentation:**
    
    Presentation result
    
    **Contribution:**
    
    Contribution result
    
    **Rating:**
    
    Rating result
    
    **Confidence:**
    
    Confidence result
    \\ \hspace*{\fill} \\
    Your final review, do not include any additional commentary:
    \end{tcolorbox}
    \caption{Prompt for the Answer Synthesizer in generating the full review at the root level.}
    \label{fig:prompt_answer_root_full}
\end{figure*}

\begin{figure*}[t]
    \centering
    \small
    \begin{tcolorbox}[colframe=black, colback=white, width=\textwidth, boxrule=0.2mm]
    You are an expert reviewer tasked with providing feedback comments for a scientific paper. You will receive the full paper content and a set of review question-answer pairs which are obtained through review process with in-depth understanding and analysis of the paper. These review Q\&A pairs will be very helpful for you to give accurate and insightful feedback comments. Please follow the instructions below:
    \\ \hspace*{\fill} \\
    INSTRUCTIONS
    
    1. You should first carefully read through the entire paper.
    
    2. It’s important to use the review questions and their corresponding answers as reference to guide and enhance your review thinking process. However, if after reading the entire paper you think some viewpoints or insights in the review Q\&A pairs to be incorrect or insufficient, please disregard these incorrect ones and refine the insufficient ones with your own expert judgment.
    
    3. Identify weak points of the paper, and write them as feedback comments. For each of your comments, it should:
    
    - Focus on the paper's weaknesses, limitations, potential flaws, and areas for improvement, or raise questions that highlight the need for clarification and further analysis.
    
    - Focus on major comments that are important and have a significant impact on the paper's quality, as opposed to minor comments about things like writing style or grammar.
    
    - Be specific and in-depth, identifying particular gaps or issues unique to this paper rather than making superficial or generic criticisms that could apply to any academic work.
    
    - Be detailed, providing comprehensive context and extensive elaboration on the identified issue, including specific aspects of the methodology, results, or claims, etc that require improvement, explaining why these issues matter, how they impact the paper's validity or contribution, what specific changes would address the concerns, ensuring substantive enough for authors to fully understand both the problem and the path to resolution.
    
    - Provide detailed evidence from the paper (e.g., quotes, section references, or data points) to support your point. For example, if a claim is unsupported, identify the exact statement and explain what evidence is missing; if a methodology is unclear, reference the section and describe what additional details are needed.
    \\ \hspace*{\fill} \\
    INPUT
    
    - Paper Content: {PAPER CONTENT}
    
    - Questions and answers: {QUESTIONS AND ANSWERS}
    \\ \hspace*{\fill} \\
    OUTPUT FORMAT
    
    Write your feedback comments as a JSON list of strings, for example: ["feedback comment1", "feedback comment2"]. 
    
    Your feedback comments, do not include any additional commentary:
    \end{tcolorbox}
    \caption{Prompt for the Answer Synthesizer in generating actionable feedback comments at the root level.}
    \label{fig:prompt_answer_root_comments}
\end{figure*}

\end{document}